\documentclass[runningheads]{llncs}

\usepackage{eccv}

\usepackage{eccvabbrv}

\usepackage{graphicx}
\usepackage{wrapfig}
\usepackage{booktabs}
\usepackage{colortbl}
\usepackage{multirow}
\usepackage{pifont}
\newcommand{\cmark}{\ding{51}}
\newcommand{\xmark}{\ding{55}}
\usepackage[accsupp]{axessibility}  

\usepackage{hyperref}

\usepackage{orcidlink}
\usepackage{pdfpages}

\begin{document}


\title{SparseCtrl-HOI: Sparse Temporal Control for Human-Object Interaction Video Generation} 

\titlerunning{SparseCtrl-HOI}

\author{Shenbo Xie \and
Mingrui Cai \and
Xu Yang \and 
Yifei Liu \and
Changxing Ding\thanks{Corresponding author.}}

\authorrunning{S.~Xie et al.}

\institute{South China University of Technology\\
\email{eeshbx34@scut.edu.cn, chxding@scut.edu.cn}
}

\maketitle

\begin{figure*}[ht]
\centering
\vspace{-5mm}
\includegraphics[width=1.0\linewidth]{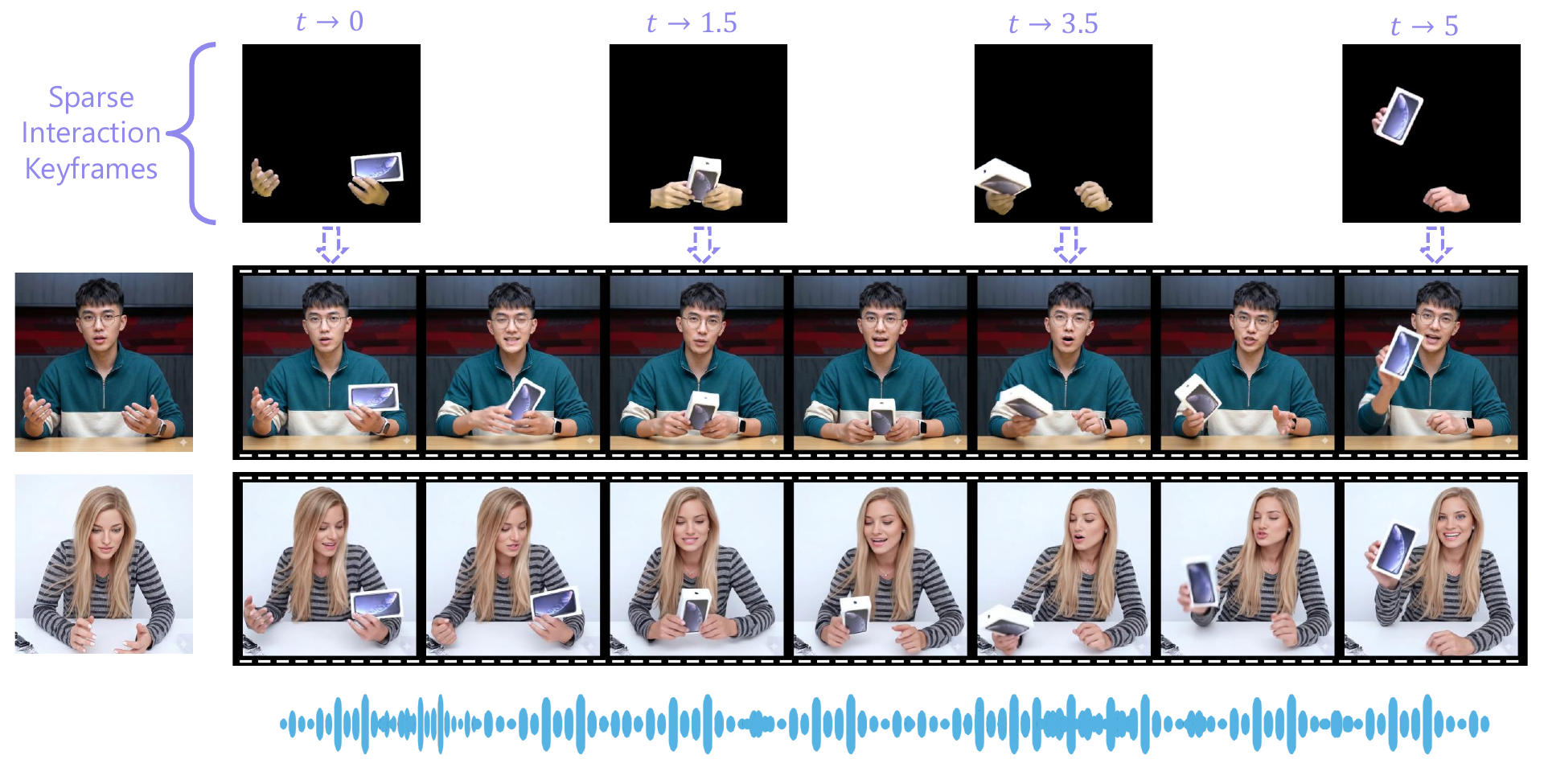}
\vspace{-1em}
\caption{Synthesized HOI video examples from our framework, demonstrating its effectiveness in sparse temporal control. For privacy, all avatar reference images are generated using Nano Banana\cite{comanici2025gemini}.}

\label{fig:teaser}
\vspace{-1em}
\end{figure*}

\begin{abstract}
Human-Object Interaction (HOI) video generation aims to synthesize realistic videos of humans manipulating diverse objects, serving as a promising avenue for AI-driven live streaming e-commerce. A primary obstacle in this domain lies in the complexity of modeling fine-grained physical dynamics and the intricate spatial-temporal coordination between human hands and objects. Existing approaches to this problem typically rely on dense temporal guidance, e.g., frame-wise hand-object pose sequences, to strictly control the interaction process. However, such dense guidance incurs high annotation costs and affects motion synthesis diversity. To overcome these limitations, we introduce SparseCtrl-HOI, a novel sparse temporal control framework for HOI video generation. It requires only a few keyframes that capture interaction states at designated timestamps. Specifically, we employ a Time-Controlled Rotary Positional Embedding (TiRoPE) mechanism to temporally anchor these keyframes while preserving their spatial integrity. Subsequently, to govern the dynamics across intermediate frames, we propose a Motion Prior Injection Module that leverages Multimodal Large Language Models (MLLMs) to extract high-level motion priors. This empowers the model to hallucinate logically and physically plausible transitions. Furthermore, we build SparseHOI-5K, a high-quality and richly annotated dataset for HOI video generation with sparse temporal control.  Comprehensive evaluations confirm that our method substantially reduces annotation overhead while synthesizing superior live-streaming e-commerce videos. Both our code and dataset are publicly available at https://mpi-lab.github.io/SparseCtrl-HOI.
\end{abstract}

\section{Introduction} 
\label{sec:intro}

Human-Object Interaction (HOI) video generation\cite{HOMA,wang2025dreamactorh1,zhang2025vhoi,liu2025byteloom,xu2026anchorcrafter} aims to synthesize realistic videos of humans interacting with diverse objects, which serves as a foundational technology for emerging applications in digital entertainment\cite{lei2024humanvideo} and virtual reality\cite{zhu2023human,sui2025surveyhumaninteractionmotion}. In the context of live streaming e-commerce, this technology holds significant potential by enabling AI-driven avatars to perform continuous, high-fidelity product demonstrations, thereby overcoming the physical and temporal constraints of human streamers. However, generating high-quality HOI videos remains challenging, as it requires modeling fine-grained physical dynamics and precise spatial-temporal coordination between hands and manipulated objects.

Recent advances in HOI video generation have introduced various control mechanisms to guide the interaction process. These approaches can be broadly divided into two groups by the granularity of their guidance signals. The first relies on dense motion priors~\cite{xu2026anchorcrafter,wang2025dreamactorh1,liu2025byteloom}. Their generation process is strictly guided by frame-wise conditions, e.g., 2D pose sequences\cite{xu2026anchorcrafter}, 3D mesh sequences~\cite{wang2025dreamactorh1}, and relative coordinate maps~\cite{liu2025byteloom}. The second explores sparse spatial control\cite{HOMA,zhang2025vhoi}. They derive complex interactions from a small amount of spatial input, e.g., sparse keypoints and object trajectories, by employing techniques like decoupled guidance\cite{HOMA} and motion densification\cite{zhang2025vhoi}. Despite this difference, both groups share a common paradigm: they depend on dense temporal guidance, which requires frame-wise synchronization between the generated videos and control signals.

However, this dense guidance paradigm has notable limitations. First, conditioning on poses at every frame overrides the generative model's inherent motion priors. This degrades the model into a mere texture renderer that produces rigid, repetitive motions. Second, any jitter or artifact in the input condition propagates directly to the output video, causing physically implausible transitions. Third, obtaining dense frame-level annotations for custom e-commerce scenarios is labor-intensive. These drawbacks motivate a sparse temporal control paradigm, where the model receives only key interaction states and autonomously synthesizes natural transitions between them.

Accordingly, we propose SparseCtrl-HOI, a framework for high-fidelity HOI video generation conditioned on only a few interaction keyframes.
We concatenate these interaction keyframes with noisy latents along the temporal dimension at the DiT\cite{peebles2023scalable} input, enabling the model to leverage self-attention to naturally propagate visual information from keyframes to synthesized frames. We further introduce Time-Controlled Rotary Positional Embedding (TiRoPE), which assigns each keyframe the temporal positional index of its target timestamp, ensuring that specified interaction states appear at their designated moments in the generated video.

Although TiRoPE ensures precise alignment at key timestamps, relying solely on positional constraints usually yield suboptimal motion in intermediate frames. To address this, we introduce a motion prior injection method to govern the motion transition between key states. Specifically, we leverage a Multimodal Large Language Model (MLLM)~\cite{wan2025} to extract high-level motion priors that encode hand–object pose change from the interaction keyframes. We then employ a Q-Former \cite{li2023blip} to compress these priors into motion guidance tokens, which are injected into the DiT via dedicated cross-attention layers. This enables the DiT to produce smooth, physically plausible transitions and ensures that the hand and object movements remain physically plausible even during significant view changes. Finally, recognizing that keyframe alignment and motion transition are complementary yet distinct objectives, we propose a decoupled training strategy that separates their learning, which fully exploits the respective strengths of the MLLM and DiT.

We also construct a high-quality Human-Object Interaction video dataset tailored for sparse temporal control, which will be publicly released.
To the best of our knowledge, this is the first work that achieves precise sparse temporal control in Human-Object Interaction video generation. By replacing dense temporal guidance with a sparse, motion prior-driven paradigm, our approach significantly reduces annotation cost while improving the naturalness of synthesized motions. Experiments show that our method produces superior live-streaming e-commerce videos with coherent object interactions, offering a practical and scalable solution for real-world applications.

\section{Related Work}
\label{sec:related}
\textbf{Human-Object Interaction Video Generation.} Existing Human-Object Interaction (HOI) video generation methods can be generally categorized into two groups based on the granularity of their guidance: dense guidance methods and sparse control methods.

The first category relies on dense motion priors or templates which provide frame-wise conditions to guide synthesis~\cite{xu2026anchorcrafter,wang2025dreamactorh1,liu2025byteloom}. 
Within this category, approaches can be further divided into full-body interaction generation and localized hand-object manipulation. For full-body HOI, recent works leverage dense signals such as 3D body mesh templates~\cite{wang2025dreamactorh1}, 2D pose sequences~\cite{xu2026anchorcrafter}, and relative coordinate maps ~\cite{liu2025byteloom} to enforce precise human-object alignment throughout the video. 
 For fine-grained hand–object interaction, methods typically rely on detailed spatial layouts or structural representations, such as contact-aware encodings \cite{pang2025manivideo,xue2024hoiswap} and layout-instructed diffusion \cite{fan2025rehold}. Despite achieving high-fidelity alignment, these dense controls often produce mechanical, repetitive motions.
Relatedly, in the broader context of character animation and editing, methods like Animate Anyone 2~\cite{hu2025animateanyone2} and MIMO~\cite{men2025mimo} adopt environment affordances and spatial decomposition to maintain frame-level structural consistency. For fine-grained hand-object interactions, methods typically rely on detailed spatial layouts or structural representations. For instance, recent frameworks~\cite{pang2025manivideo,xue2024hoiswap} utilize structure-aware or contact-aware representations, while Re-HOLD~\cite{fan2025rehold} employs an adaptive layout-instructed diffusion model for generalizable grasping. Despite achieving high-fidelity alignment, these dense controls often produce mechanical, repetitive motions.

To mitigate the heavy annotation burden of dense guidance, the second category explores sparse control mechanisms that derive complex interactions from minimal user inputs. These methods typically employ trajectory-guided generation. Works like HOMA ~\cite{HOMA} and VHOI ~\cite{zhang2025vhoi} synthesize full-body HOI videos from sparse keypoints or object trajectories. They achieve this by employing decoupled motion guidance or motion densification techniques to produce intermediate movements between sparse inputs. Despite reducing input complexity, these methods still convert sparse cues into dense frame-level constraints, leaving the generated interactions vulnerable to interpolation artifacts and unnatural jitters. In contrast, our method establishes a genuinely sparse, keyframe-driven paradigm that leverages MLLM-derived motion priors and specialized temporal embeddings to synthesize smooth, physically plausible transitions.

\noindent \textbf{Audio-Driven Human Video Generation.} This task aims to synthesize realistic videos of a speaking person that are not only temporally coherent with the input audio but also visually consistent with a given reference image. Existing approaches can be divided into two categories: two-stage methods and end-to-end methods. The two-stage methods first predict intermediate motion representations, such as 2D body keypoints\cite{hogue2024diffted,he2024s2gmmdiffusion,yang2025demo} or coefficients of 3D body models\cite{corona2025vlogger}, and then render video frames conditioned on these representations. However, errors in these intermediate representations may propagate to the second stage, resulting in artifacts in the synthesized videos.

Recent works thus turn to the end-to-end paradigm that directly synthesizes videos from the audio. These methods advance in several complementary directions. One line of works designs richer audio representations for tighter lip synchronization and audio-motion alignment\cite{gan2025omniavatar,sun2026coshdit}. Another line enhances fine-grained details in critical regions such as hands and faces through region-specific attention~\cite{lin2025cyberhost} or auxiliary 2D keypoints inputs~\cite{meng2025echomimicv2}. There are also works that incorporate 3D structural priors to regularize body geometry and refine local details through cascaded refinement~\cite{guan2025audcast,sun2026coshdit}. Finally, post-training preference optimization is explored to jointly improve motion naturalness and visual fidelity~\cite{liang2025alignhuman}. In this work, we build our model on the end-to-end paradigm and focus on producing realistic human-object interaction dynamics.

\section{Method}
\label{sec:method}
\begin{figure*}[t]
\centering
\includegraphics[width=1.0\linewidth]{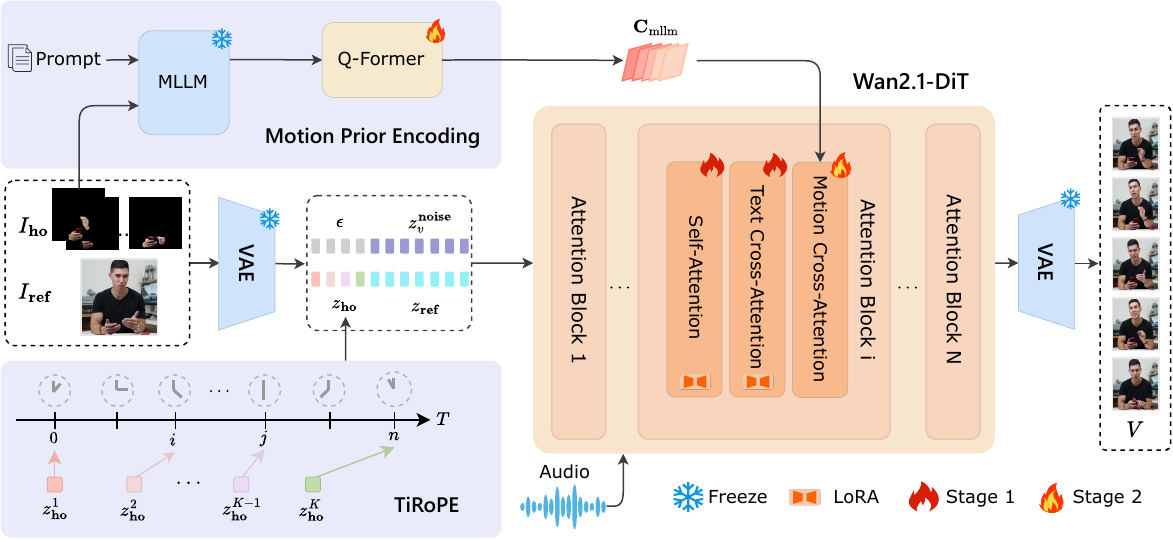}
\vspace{-2em}
\caption{The overall framework of our SparseCtrl-HOI. We adopt a two-stage training strategy built upon the Wan2.1-DiT backbone. Given a reference image $I_{\mathrm{ref}}$, an audio sequence, and sparse interaction keyframes $I_{\mathrm{ho}}^1,\ldots,I_{\mathrm{ho}}^K$ at specified timestamps, the pipeline operates as follows. Motion Prior Encoding (Top): A frozen Qwen2.5-VL~\cite{qwen2.5-VL} extracts high-level motion semantics from the prompt and keyframes. These are then compressed by a Q-Former into motion prior tokens $\mathbf{C}_{\mathrm{mllm}}$. Main Pipeline: Visual inputs are encoded by a frozen VAE~\cite{wan2025}. We concatenate these latents to $\mathbf{z}_{\mathrm{in}}$. Furthermore, the TiRoPE mechanism is applied to $\mathbf{z}_{\mathrm{ho}}$. Within the DiT blocks, we inject LoRA~\cite{hu2022lora} into the Self-Attention and Text Cross-Attention layers (Stage 1), and introduce a novel Motion Cross-Attention layer (Stage 2) in each block to integrate the $\mathbf{C}_{\mathrm{mllm}}$ tokens, enabling coherent human-object interaction video generation.}

\label{fig:framework}
\vspace{-1em}
\end{figure*}

As illustrated in Figure \ref{fig:framework}, our method takes three inputs: a reference image specifying the person's identity, an audio sequence, and a sparse set of interaction keyframes. In the following, we first introduce the preliminary formulation and our base generation pipeline in Section \ref{subsec:preliminary_pipeline}. We then detail our core innovations: Time-Controlled Rotary Positional Embedding (TiRoPE) for precise temporal anchoring (Section \ref{subsec:tirope}), the Motion Prior Injection Module for synthesizing natural intermediate dynamics (Section \ref{subsec:motion_prior}), and our Decoupled Training Strategy (Section \ref{subsec:training}). Finally, we elaborate on the curation pipeline of our HOI video dataset in Section \ref{subsec:data_curation}.

\subsection{Preliminaries and Baseline}
\label{subsec:preliminary_pipeline}
Our base architecture is based on the continuous-time flow matching paradigm formulated in OmniAvatar~\cite{gan2025omniavatar}. An input raw video $x_0 \in \mathbb{R}^{T \times H \times W \times 3}$ is compressed into a latent representation $\mathbf{z}_0 \in \mathbb{R}^{\frac{T}{4} \times \frac{H}{8} \times \frac{W}{8} \times C}$ through a pre-trained causal 3D VAE encoder $\mathcal{E}$. The generative process is governed by an Ordinary Differential Equation (ODE) \cite{lipman2022flow}, where a Diffusion Transformer (DiT) \cite{peebles2023scalable} $v_\theta$ predicts the velocity field $d\mathbf{z}_t / dt = v_\theta(\mathbf{z}_t, t, c)$. Adopting a linear interpolation path $\mathbf{z}_t = (1-t)\mathbf{z}_0 + t\boldsymbol{\epsilon}$ with noise $\boldsymbol{\epsilon} \sim \mathcal{N}(0, \mathbf{I})$, the model is optimized using the flow matching objective:
\begin{equation}
\mathcal{L}_{\text{flow}}(\theta) = \mathbb{E}_{t \sim \mathcal{U}[0,1], \mathbf{z}_0 \sim p_{\text{data}}, \boldsymbol{\epsilon} \sim p_{\text{noise}}} \left[ \left\| v_\theta(\mathbf{z}_t, t, c) - (\boldsymbol{\epsilon} - \mathbf{z}_0) \right\|_2^2 \right],
\end{equation}
where $c$ represents multimodal conditioning signals. Audio signals are integrated via a pixel-wise multi-hierarchical embedding strategy~\cite{gan2025omniavatar}.

To construct our baseline, we concatenate a set of latent conditions as inputs to DiT. Specifically, the inputs comprise the latent noisy video feature $\mathbf{z}_{\mathrm{v}}^{\mathrm{noise}} \in \mathbb{R}^{\frac{T}{4} \times \frac{H}{8} \times \frac{W}{8} \times C}$, the latent reference image feature $\mathbf{z}_{\mathrm{ref}} \in \mathbb{R}^{1 \times \frac{H}{8} \times \frac{W}{8} \times C}$ representing the person’s identity, and the latent interaction keyframe features $\mathbf{z}_{\mathrm{ho}} \in \mathbb{R}^{K \times \frac{H}{8} \times \frac{W}{8} \times C}$ for timestamps $t_1, \dots, t_K$. To align their channel dimension, we pad $\mathbf{z}_{\mathrm{ho}}$ with a random noise tensor $\boldsymbol{\epsilon}_{\mathrm{ho}}$. The above aggregated inputs can be formulated as:
\begin{equation}
    \mathbf{z}_{\mathrm{in}} = \text{Concat}_T \left( \text{Concat}_C(\mathbf{z}_{\mathrm{v}}^{\mathrm{noise}}, \mathbf{z}_{\mathrm{ref}}), \text{Concat}_C(\boldsymbol{\epsilon}_{\mathrm{ho}}, \mathbf{z}_{\mathrm{ho}}) \right),
\end{equation}
where $\text{Concat}_C$ and $\text{Concat}_T$ represent concatenation along the channel and temporal dimensions, respectively. This design leverages the inherent self-attention operations within DiT to transfer the human-object interaction poses and object textures contained in $\mathbf{z}_{\mathrm{ho}}$ to $\mathbf{z}_{\mathrm{v}}^{\mathrm{noise}}$. Finally, similar to other diffusion-based video generation models~\cite{gan2025omniavatar,wang2025dreamactorh1,HOMA}, this baseline imposes the standard Rotary Positional Embedding (RoPE) \cite{su2023roformerenhancedtransformerrotary} on the input of the DiT model. 

Moreover, we apply hand-object masks to remove background interference in the interaction keyframes. To facilitate smooth blending between the reserved hand-object region and the surrounding background, we refine the masks with morphological operations and Gaussian filtering. Finally, we apply data augmentations, including HSV jittering and Scaling, to the hand regions, which enables the DiT model to focus on the hand poses rather than the appearances from the keyframes.

\subsection{Time-Controlled Rotary Positional Embedding}
\label{subsec:tirope}
The baseline described above lacks temporal control over when a specific interaction pose occurs. To address this, we introduce Time-Controlled Rotary Positional Embedding (TiRoPE). First, we impose standard RoPE on $\mathbf{z}_{\mathrm{v}}^{\mathrm{noise}}$. Second, for the latent feature $\mathbf{z}_{\mathrm{ho}}^i$ of the $i$-th interaction keyframe, we simply assign its target timestamp as the temporal positional index for RoPE. This gives the keyframe and its target video frame identical temporal embeddings, thereby encouraging high attention scores between them during DiT self-attention. We keep the spatial positional index unchanged to maintain spatial consistency with the interaction keyframes. 

\subsection{Motion Prior Injection}
\label{subsec:motion_prior}

While TiRoPE enables effective HOI pose and object texture transfer at specified timestamps, it provides no explicit guidance for the other frames. This results in unnatural transitions between two nearby interaction keyframes. This is particularly severe when the interaction pose undergoes significant changes.

To bridge sparse control signals with physically plausible intermediate dynamics, we leverage the semantic reasoning ability of Multimodal Large Language Models (MLLMs). Specifically, we feed the interaction keyframes $I_{\mathrm{ho}}^1, \dots, I_{\mathrm{ho}}^K$ into Qwen2.5-VL \cite{qwen2.5-VL} along with a task-specific prompt to extract high-level motion priors that describe the pose changes between keyframes:

\begin{quote}

\textit{"We are working on the hand-object interaction video generation task, where one person is introducing a product using hands in each generated video. To facilitate generation, we have provided four keyframes in the temporal order. We want you to infer the complete hand-object interaction process based on the four keyframes. In particular, we want you to infer the change of hand pose, object pose, the hand-object contact region in this process as priors for the video generation."} 

\end{quote}

We adopt tokens produced by the final MLLM layer as raw motion priors $\mathbf{R}_{\mathrm{mllm}}$. To compress these lengthy representations, we introduce a Q-Former-based architecture\cite{li2023blip}. It includes four cascaded modules, each of which contains one self-attention layer, one cross-attention layer, and one feed-forward network. To facilitate the subsequent coordination with DiT, we first project the token dimension in $\mathbf{R}_{\mathrm{mllm}}$ to the DiT's embeddings via an MLP. We then employ 64 learnable queries $\mathbf{Q}$ to extract condensed motion priors via cross-attention from $\mathbf{R}_{\mathrm{mllm}}$: 
\begin{equation}
\mathbf{C}_{\mathrm{mllm}} = \text{Q-Former}(\mathbf{Q}, \text{MLP}(\mathbf{R}_{\mathrm{mllm}})).
\end{equation}
Finally, we inject $\mathbf{C}_{\mathrm{mllm}}$ into the DiT backbone through dedicated cross-attention layers in each transformer block. $\mathbf{C}_{\mathrm{mllm}}$ then guides DiT to hallucinate logical, smooth motion transitions between sparse interaction keyframes. Detailed tensor construction, TiRoPE index mapping, and motion cross-attention formulation are provided in the supplementary material.

\subsection{Decoupled Training}
\label{subsec:training}
\indent MLLMs excel at abstract reasoning, while generative models target concrete visual synthesis. They essentially handle features of different purposes and granularities. To avoid their feature entanglement, we propose a decoupled, two-stage training strategy:

\noindent \textbf{Stage 1: Appearance Feature Transfer.} This stage teaches the DiT model to extract the subject and object appearance features from the reference image and interaction keyframes, respectively. It also enables the model to adhere strictly to the interaction poses dictated by the TiRoPE-anchored keyframes. We unfreeze the patch embedding and the audio pack~\cite{gan2025omniavatar}, and employ Low-Rank Adaptation (LoRA)~\cite{hu2022lora} on the pre-trained DiT blocks.

\noindent \textbf{Stage 2: Motion Transition Inference.} This stage envisions physically plausible motion transitions between two nearby interaction keyframes. We freeze all LoRA parameters optimized in Stage 1, thus preserving the visual synthesis ability of the DiT model. We then exclusively train the newly introduced Q-Former alongside cross-attention layers inserted to the DiT model. These new layers of cross-attention acquire motion cues from $\mathbf{C}_{\mathrm{mllm}}$ and enable the DiT model to synthesize temporally coherent movements of the hand and the object. 

In the experimentation section, we justify that this decoupled training strategy achieves better performance than naïve one-stage training.

\begin{table}[t]
\centering
\caption{Comparisons between SparseHOI-5K and existing public HOI datasets. ``-'' indicates that the value is not reported in the official release. \emph{Hand Cut}/\emph{Obj.\ Cut} denote cropped hand/object regions; \emph{Obj-free} denotes object-removed videos via inpainting.}
\vspace{-1em}
\label{tab:dataset_comparison}
\renewcommand{\arraystretch}{1.1}
\resizebox{\linewidth}{!}{%
\begin{tabular}{lcccccccc}
\toprule
Dataset & Videos & Dur.\ (h) & Actor & Objects & Audio & Hand Cut & Obj.\ Cut & Obj-free \\
\midrule
HOIGen-1M~\cite{HOIGen} & 1M & 2.2k & - & 15,000+ & \xmark & \xmark & \xmark & \xmark \\
AnchorCrafter-400~\cite{xu2026anchorcrafter} & 356 & 2 & 11 & 286 & \xmark & \cmark & \cmark & \xmark \\
\textbf{SparseHOI-5K} & 4,850 & 10 & 34 & 1,000+ & \cmark & \cmark & \cmark & \cmark \\
\bottomrule
\end{tabular}%
}
\end{table}

\begin{figure*}[t]
\centering
\includegraphics[width=1.0\linewidth]{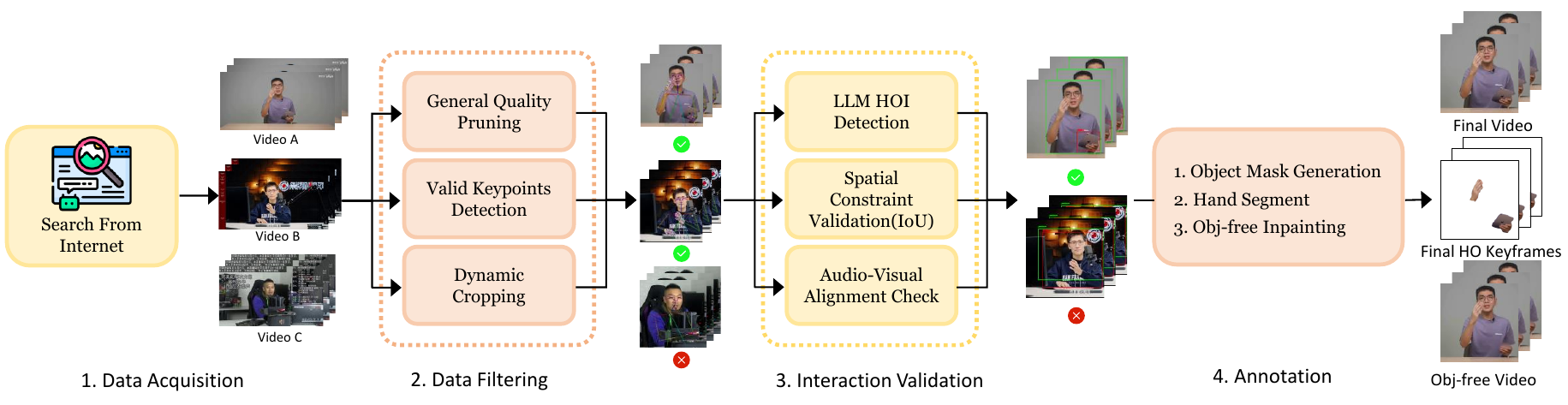}
\vspace{-2em}
\caption{An overview of SparseHOI-5K dataset collection pipeline.}
\label{fig:datapipeline}
\vspace{-1em}
\end{figure*}

\subsection{Dataset Construction Pipeline}
\label{subsec:data_curation}
Although HOIGen-1M~\cite{HOIGen} offers million-scale text-to-video HOI clips, no public dataset targets audio-driven live-commerce HOI video generation with sparse temporal control, paired audio, hand/object masks, and object-free references.
To bridge this gap, we build a high-quality HOI video dataset named SparseHOI-5K that includes 4,850 clips and make it publicly available. It encompasses 34 human identities and over 1000 promoted commodities in live streaming e-commerce scenarios. All videos are unified to a resolution of 512$\times$512 pixels at 25 FPS, with synchronized audio sampled at 16 kHz. Moreover, we provide rich multi-modal annotations for each clip, including hand masks, object masks, and inpainted videos frames where the commodities are removed.
As summarized in Table~\ref{tab:dataset_comparison}, SparseHOI-5K offers richer multi-modal annotations than existing public HOI datasets, which is essential for our sparse keyframe-conditioned setting.
In the following, we introduce our rigorous, four-stage data processing pipeline as depicted in Figure~\ref{fig:datapipeline}:

\noindent\textbf{Phase I: Data Acquisition.} We download raw videos from two live-streaming e-commerce platforms, YouTube\cite{youtube} and Bilibili\cite{bilibili}, using the Google Data API and open-source downloaders\cite{youtube_data_api,yt-dlp}. We transcode all raw videos to a unified H.264 codec\cite{wiegand2003overview}. We also apply a scene detection algorithm~\cite{pyscenedetect} to divide each video into segments based on visual continuities. Finally, we sample fixed-length clips of 10 seconds within each segment.

\noindent\textbf{Phase II: Data Filtering.} We further prune low-quality clips based on static content detection~\cite{opencv_library} and esthetic scores~\cite{emon2020novel}. We also utilize DWpose~\cite{yang2023dwpose} to detect body bounding boxes and keypoints, by which we keep only single-person sequences. Finally, we center each subject and minimize extraneous background in the resized video frames based on the location of human bounding boxes. 

\noindent\textbf{Phase III: Interaction Validation.} We leverage Qwen2.5-VL~\cite{qwen2.5-VL} to validate that each clip contains exactly one person interacting with an object, simultaneously estimating the bounding boxes of the target objects. To handle MLLM hallucination, we calculate the Intersection over Union (IoU) between bounding boxes of the subject and the object; clips with excessive IoU—indicating the model may misidentify the subject as the object—are excluded. Finally, we employ SyncNet~\cite{Chung16syncnet} to ensure lip-sync consistency, and discard clips with audio-visual mismatches.

\noindent\textbf{Phase IV: Fine-Grained Annotation.} We utilize SAM2~\cite{ravi2024sam2} for object segmentation. We adopt a cascaded MediaPipe\cite{lugaresi2019mediapipe} and SAM\cite{kirillov2023segment} approach for precise hand segmentation. These steps enable us to obtain hand-object regions in all video frames and therefore facilitate clean interaction keyframe production in Section~\ref{subsec:motion_prior}. Finally, we employ an image inpainting tool named ProPainter~\cite{zhou2023propainter} to synthesize object-free versions of all video clips. In this way, we can randomly select one inpainted frame as the reference image of the subject.

\section{Experiments}
\label{sec:experiments}

\subsection{Datasets and Metrics}
\label{subsec:datasets_metrics}

\noindent \textbf{Datasets.} We partition our SparseHOI-5K dataset into mutually exclusive training and testing sets based on subject identities. They cover 4,800 clips and 50 clips, respectively. This identity-disjoint split ensures that there is no data leakage during training, allowing us to accurately measure the model's performance on unseen individuals. Furthermore, to assess the model's cross-dataset generalization capability, we directly evaluate its performance on the AnchorCrafter testing set\cite{xu2026anchorcrafter}. The testing set includes 8 celebrity images and 10 source videos in which four persons interact with four simple objects.

\noindent \textbf{Evaluation Metrics.} To comprehensively assess the quality of the generated videos, we employ a combination of standard video generation metrics and customized protocols tailored for sparse temporal control Human-Object Interactions (HOI). For standard metrics, we measure overall spatiotemporal consistency and distribution similarity using Fréchet Image Distance (FID), Fréchet Video Distance (FVD)~\cite{unterthiner2019fvd} and VBench~\cite{huang2024vbench}, which includes specifically Motion Smoothness (MS), Temporal Flickering (TF) and Aesthetic Quality (AQ). For audio-visual alignment, we compute the Sync-C score using SyncNet~\cite{Chung16syncnet}. 

Moreover, we introduce the following three customized metrics to evaluate the motion transition naturalness, control precision by interaction keyframes and the quality of human-object interaction:

\noindent \textbf{MS-RAFT:} To rigorously quantify the temporal smoothness of the generated videos at the pixel-level, we introduce the MS-RAFT warping error. Specifically, we use a pre-trained RAFT-Large model\cite{teed2020raft} to estimate the backward optical flow $F_{t \to t-1}$ from frame $I_t$ to frame $I_{t-1}$. We then warp $I_{t-1}$ to the spatial coordinates of $I_t$ by bilinear interpolation guided by $F_{t \to t-1}$. We compute the Mean Absolute Error (MAE) between the warped frame and the raw frame. The resulting MS-RAFT score is averaged across all adjacent frames in the sequence, where a lower value signifies superior temporal continuity.

\noindent \textbf{Ti-SSIM:} To evaluate the alignment of HOI poses at designated control timestamps, we introduce the timestamp structural similarity (Ti-SSIM). We compute the SSIM score~\cite{wang2004ssim} between the ground-truth reference frame and the generated frame in designated timestamp with a local temporal window (e.g., $\pm 10$ frames). We define the interaction accuracy score as the maximum SSIM value within this window and subsequently compute the mean score across all generated videos. Since the previous method did not control the interactions at sparse timestamps, we only adopt this metric in the ablation experiments.

\noindent \textbf{HOI-VLM:} To evaluate the complex dynamics between humans and objects, we leverage Qwen2.5-VL to automatically score the physical plausibility and diversity of the interaction. Specifically, we uniformly sample 20 frames from the generated video and provide them as input to Qwen2.5-VL. We instruct the VLM to score the video (from 1 to 5) across two essential dimensions: \textit{Physical Plausibility} (e.g., absence of interpenetration or unnatural levitation) and \textit{Interaction Diversity} (e.g., rich kinematic variations rather than rigid stasis). Higher scores indicate better human-object interactions.

\subsection{Experimental Settings}
\label{subsec:experimental_settings}

\noindent \textbf{Implementation Details.} 
Our framework is implemented in PyTorch. Following the decoupled strategy in Section~\ref{subsec:training}, training proceeds in two stages on the full training set. Both stages use a learning rate of  $5 \times 10^{-6}$, a global batch size of 8, and 4 hand-object reference keyframes per video. Stage 1 runs for 15,000 steps to learn appearance feature transfer, and Stage 2 continues for 10,000 steps to refine motion transitions.

\noindent \textbf{Baselines.} We compare the performance of SparseCtrl-HOI with the following four state-of-the-art, open-source models relevant to our task:
\begin{itemize}
    \item \textbf{AnchorCrafter~\cite{xu2026anchorcrafter}} is a dense-control method that uses human pose, hand mesh maps, and object depth maps for HOI video generation. We extract these control signals following its original pipeline. Since it requires object images at ±45° viewpoints, which are not available in our setting, we duplicate the image of a single available object three times as a substitute.
    \item \textbf{OmniAvatar~\cite{gan2025omniavatar}} is an audio-driven framework that synthesizes talking avatar videos from a single image. As it lacks native HOI support, we adapt it by conditioning it on the first frame of each raw test clip. We also build \textbf{OmniAvatar-BL}, which takes the same inputs (reference image, four keyframes, and audio) but uses only the baseline condition injection with standard 3D RoPE, excluding TiRoPE and Motion Prior Injection.
    \item \textbf{VACE~\cite{jiang2025vace}} is a unified framework for video creation and editing through multimodal conditions, and \textbf{Phantom~\cite{liu2025phantom}} is a subject-consistent video generation model that focuses on identity preservation. We provide both the reference identity image and one inpainted object image for the two models to synthesize the output.
\end{itemize}

To facilitate fair comparison, we use the 14B version of OmniAvatar, VACE, and Phantom models throughout this paper.

\begin{figure}[p] 
  \centering
  
  \begin{subfigure}{\linewidth} 
    \centering
    \includegraphics[width=1\linewidth]{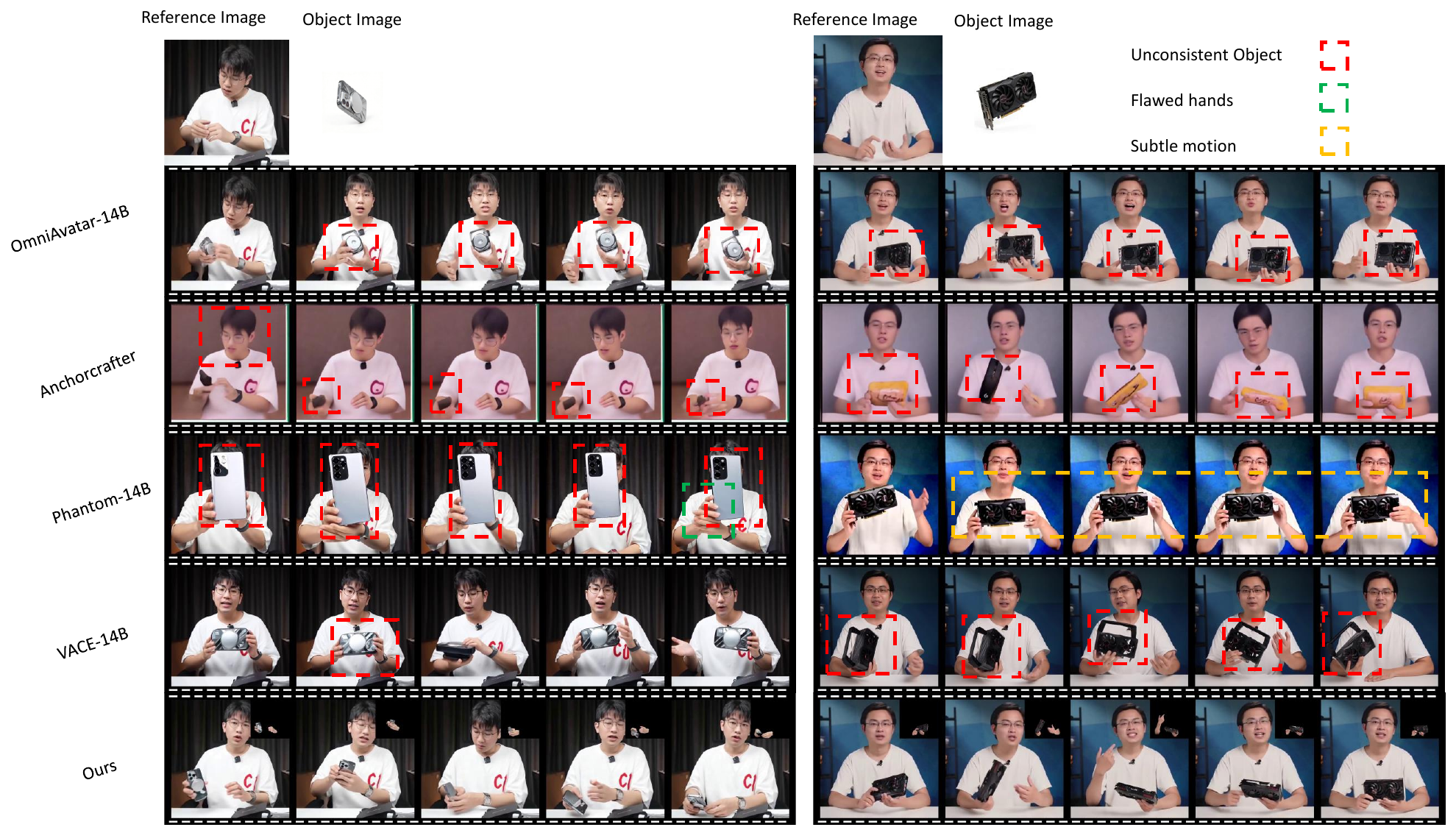} 
    \caption{Comparisons on SparseHOI-5K dataset.}
    \label{fig:exp_our}
  \end{subfigure}
  
  \vspace{0.3cm} 
  
  \begin{subfigure}{\linewidth}
    \centering
    \includegraphics[width=1.0\linewidth]{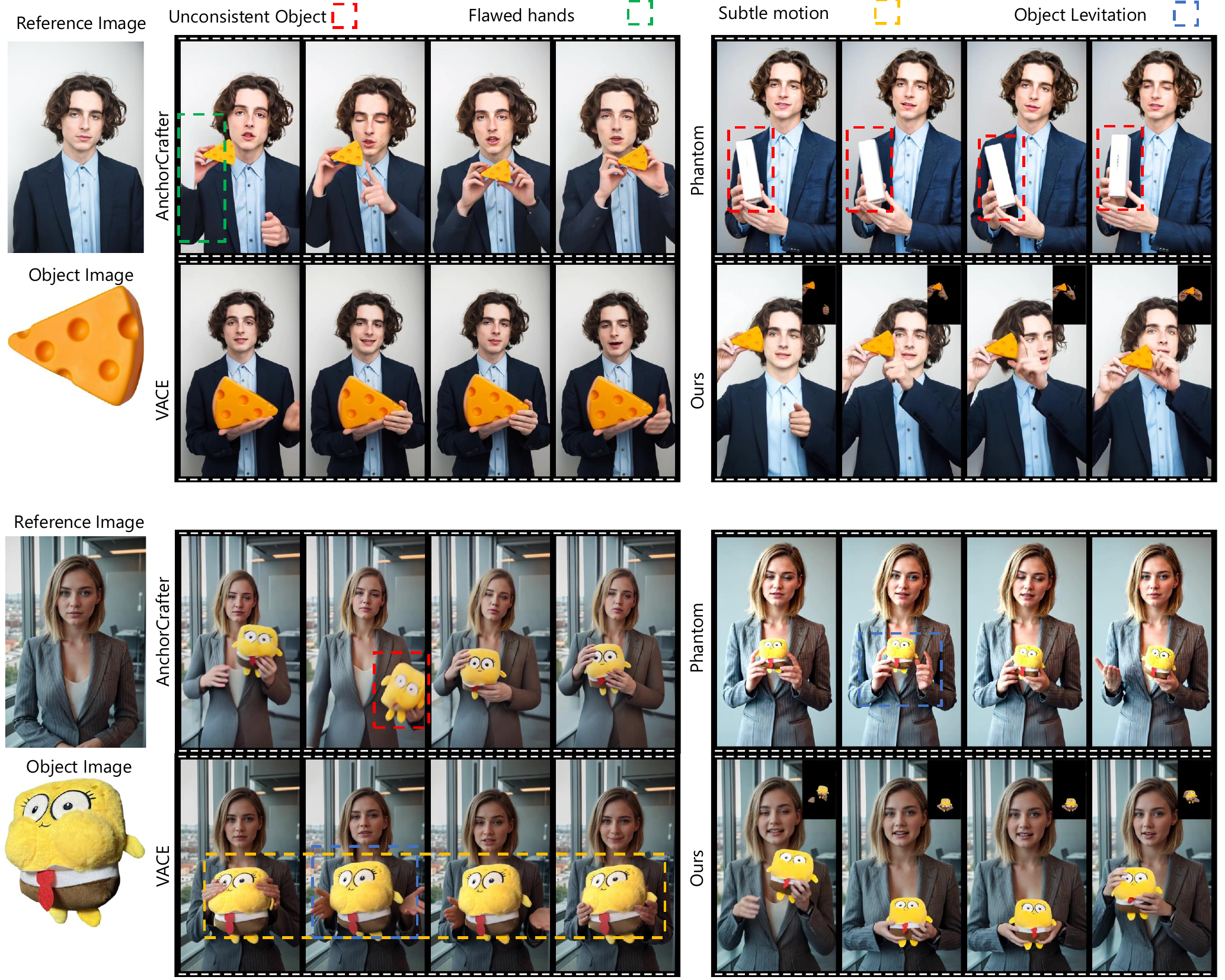} 
    \caption{Comparisons on AnchorCrafter dataset.}
    \label{fig:exp_b}
  \end{subfigure}
  
  \caption{Overall qualitative comparison.}
  \label{fig:overall_comparison}
\end{figure}

\subsection{Qualitative Comparisons}
\label{subsec:qualitative}
Figure~\ref{fig:overall_comparison} presents qualitative comparisons between our SparseCtrl-HOI and state-of-the-art methods. As highlighted by the red bounding boxes, existing methods struggle to maintain object consistency during dynamic interactions. When interaction motions change, they often produce severe blurring or hallucinate entirely different objects. In the AnchorCrafter dataset~\cite{xu2026anchorcrafter}, AnchorCrafter causes hand artifacts whenever the given reference pose is incorrect. However, in the SparseHOI-5K dataset, AnchorCrafter further fails to preserve subject identity and tends to generate globally blurry frames. We attribute the differing margins across the two test sets to data distribution: AnchorCrafter is trained and tested on simpler data, but generalizes poorly to SparseHOI-5K's more diverse identities, commodities, and hand-object motions. Furthermore, We still outperform it on its own test set using four keyframes instead of dense frame-wise controls. OmniAvatar~\cite{gan2025omniavatar} achieves reasonable realism in human-object interactions, yet its generated poses remain constrained by the reference image and cannot synthesize large-scale pose variations. Phantom~\cite{liu2025phantom} and VACE~\cite{jiang2025vace}, in turn, tend to produce physically implausible results such as floating or oversized objects. In contrast, our method significantly reduces these artifacts, while enabling precise control over interactions at specified timestamps. Additional video demonstrations are provided in the supplementary material.

\begin{table*}[ht]
\centering
\small
\caption{Quantitative comparisons between SparseCtrl-HOI and state-of-the-art methods on our SparseHOI-5K dataset and AnchorCrafter dataset. $\uparrow$ indicates higher is better and $\downarrow$ indicates lower is better. The best results are highlighted in \textbf{bold}.}
\label{tab:main_results}
\renewcommand{\arraystretch}{1.2}
\resizebox{\textwidth}{!}{%
\begin{tabular}{ll ccccc cc cccc}
\toprule
Dataset & Method & FID$\downarrow$ & FVD$\downarrow$ & MS$\uparrow$ & TF$\uparrow$ & AQ$\uparrow$ & Sync-C$\uparrow$ & MS-RAFT$\downarrow$ & HOI-VLM$\uparrow$ \\
\midrule
\multirow{4}{*}{AnchorCrafter}
& AnchorCrafter~\cite{xu2026anchorcrafter} & 91.9 & 726.4 & \textbf{0.99} & 0.98 & 0.46 & - & 0.43 & 2.90 \\
& VACE~\cite{jiang2025vace}               & 249.0 & 3809.2 & 0.98 & 0.97 & 0.418 & - & 0.60 & 2.90 \\
& Phantom~\cite{liu2025phantom}           & 254.3 & 2544.3 & 0.96 & 0.98 & 0.43 & - & 0.53 & 2.50 \\
& \textbf{SparseCtrl-HOI}                 & \textbf{91.4} & \textbf{725.8} & \textbf{0.99} & \textbf{0.99} & \textbf{0.48} & - & \textbf{0.27} & \textbf{3.00} \\
\midrule
\multirow{4}{*}{SparseHOI-5K}
& AnchorCrafter~\cite{xu2026anchorcrafter} & 258.4 & 2859.8 & \textbf{0.99} & 0.98 & 0.43 & - & 0.79 & 1.18 \\
& OmniAvatar~\cite{gan2025omniavatar} & 131.2 & 1207.0 & \textbf{0.99} & \textbf{0.99} & 0.43 & \textbf{7.04} & 0.47 & 2.76 \\
& OmniAvatar-BL & 129.3 & 1110.9 & 0.98 & 0.98 & 0.43 & 6.99 & 0.49 & 2.78 \\
& VACE~\cite{jiang2025vace} & 212.9 & 1613.3 & 0.98 & 0.98 & 0.43 & - & 0.65 & 2.86 \\
& Phantom~\cite{liu2025phantom} & 254.2 & 2701.7 & \textbf{0.99} & 0.98 & \textbf{0.44} & - & 0.48 & 1.56 \\
& \textbf{SparseCtrl-HOI} & \textbf{89.1} & \textbf{629.6} & \textbf{0.99} & \textbf{0.99} & 0.43 & 6.98 & \textbf{0.45} & \textbf{2.92} \\
\bottomrule
\end{tabular}%
}
\end{table*}

\begin{table}[ht]
\centering
\caption{Ablation study on key architectural component, condition injection scheme, and inference number of sparse keyframes $K$ on SparseHOI-5K. Lower is better for FID, FVD, and MS-RAFT; higher is better for Ti-SSIM.}
\label{tab:ablation_compact}
\scriptsize
\setlength{\tabcolsep}{3.2pt}
\renewcommand{\arraystretch}{0.95}
\resizebox{\linewidth}{!}{%
\begin{tabular}{llcccc}
\toprule
Group & Variant & FID$\downarrow$ & FVD$\downarrow$ & Ti-SSIM$\uparrow$ & MS-RAFT$\downarrow$ \\
\midrule
\multirow{4}{*}{Component}
& w/o Data Aug. & 95.14 & 709.14 & 0.65 & 0.52 \\
& w/o TiRoPE & 99.28 & 697.84 & 0.52 & 0.64 \\
& w/o M.P.I. & 90.86 & 651.18 & \textbf{0.66} & 0.53 \\
& Full model & \textbf{89.17} & \textbf{629.61} & \textbf{0.66} & \textbf{0.45} \\
\midrule
\multirow{3}{*}{Injection}
& Reference along temporal axis & 111.81 & 822.67 & 0.60 & 0.66 \\
& Keyframes along channel axis & 202.54 & 1292.19 & 0.50 & 0.79 \\
& Full model & \textbf{89.17} & \textbf{629.61} & \textbf{0.66} & \textbf{0.45} \\
\midrule
\multirow{5}{*}{$K$ at inference}
& 2 & 115.64 & 795.32 & \textbf{0.67} & 0.45 \\
& 3 & 102.25 & 664.77 & \textbf{0.67} & \textbf{0.44} \\
& 4 & 89.17 & 629.61 & 0.66 & 0.45 \\
& 6 & 82.20 & 544.05 & 0.66 & 0.48 \\
& 8 & \textbf{81.34} & \textbf{515.84} & \textbf{0.67} & 0.46 \\
\bottomrule
\end{tabular}%
}
\end{table}

\subsection{Quantitative Comparisons}
\label{subsec:quantitative}
Table \ref{tab:main_results} presents quantitative comparisons. In the AnchorCrafter dataset, our model demonstrates the best performance across all metrics, indicating the superiority of our method on the simple human object interaction dataset. In the SparseHOI-5K dataset, our method achieves the best FID and FVD scores, indicating superior visual fidelity. In contrast, baseline models show notable degradation on these metrics, largely due to poor preservation of subject identity and object texture. Our VBench scores are competitive with all baselines, confirming strong object consistency and temporal smoothness. Notably, OmniAvatar achieves a slightly higher Sync-C score (7.04 vs. our 6.98), as it conditions solely 
on the first frame for speech alignment, whereas our framework must jointly balance audio-visual synchronization with complex hand-object interactions. Finally, our method attains state-of-the-art results on both MS-RAFT and HOI-VLM, demonstrating clear advantages in physical plausibility, transition naturalness, and interaction diversity. We further conduct a user study on visual plausibility and motion diversity, where our method receives the highest user preference; the results are reported in the supplementary material.

\begin{figure*}[t]
\centering
\includegraphics[width=0.55\textwidth]{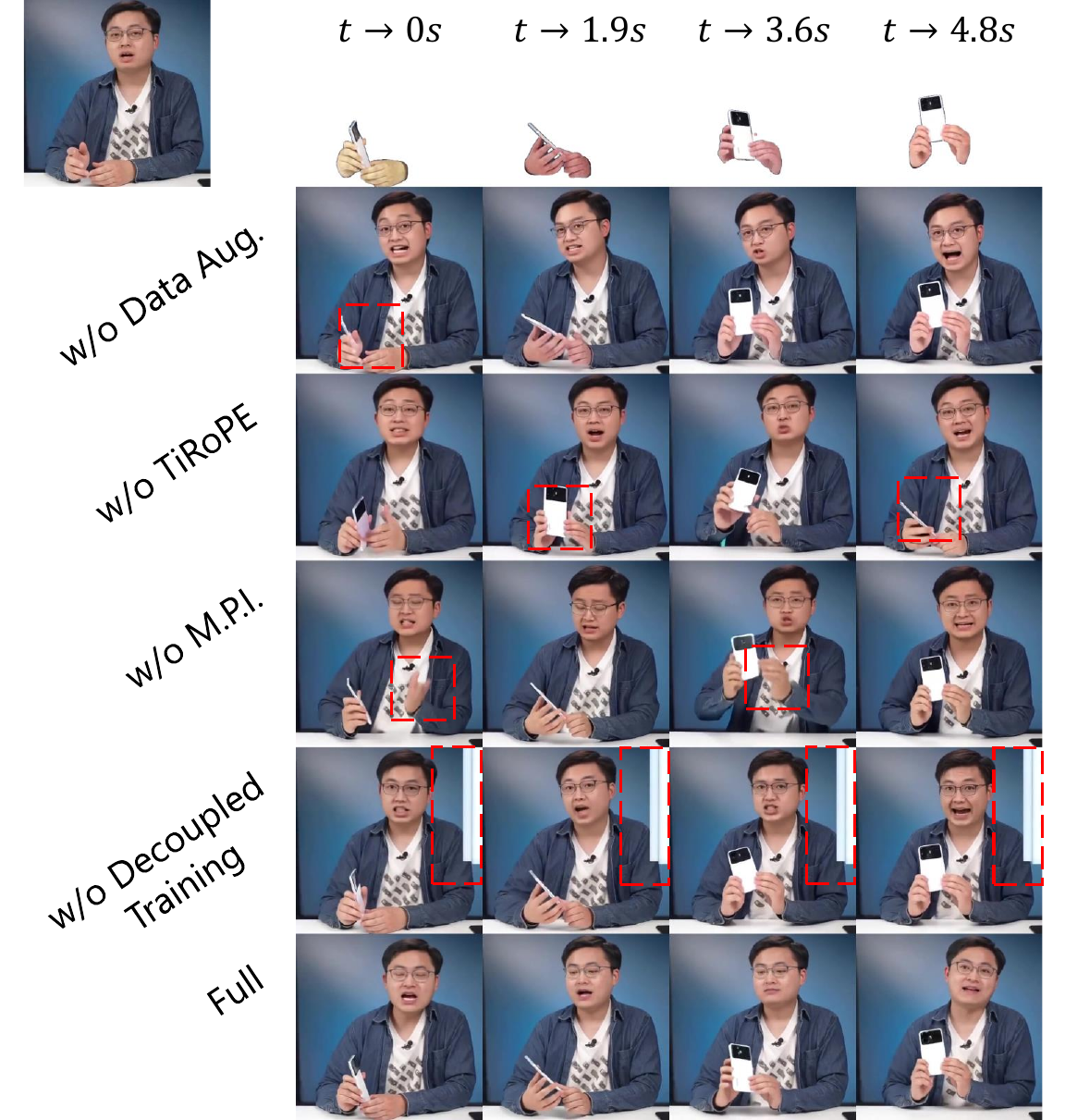}
\caption{Qualitative ablations.}
\label{fig:ablation_qual}
\end{figure*}

\subsection{Ablation Study}
\label{subsec:ablation}

We perform ablation studies on SparseHOI-5K to justify the contribution of each proposed component. The results are detailed in Table~\ref{tab:ablation_compact}.

\noindent \textbf{Impact of Data Augmentation.} 
We remove the HSV jittering and morphological operations introduced in Section~\ref{subsec:preliminary_pipeline}. As shown in the first row of Table~\ref{tab:ablation_compact}, both FID and FVD scores significantly degrade. Qualitatively, the generated videos exhibit visible cut-and-paste artifacts (Figure~\ref{fig:ablation_qual}), confirming that data augmentation is essential for smooth texture fusion.

\noindent \textbf{Design of Condition Injection.}
The Injection group in Table~\ref{tab:ablation_compact} validates our asymmetric injection design. In our full model, the reference latent is broadcast along time and concatenated channel-wise so that it conditions every frame, while sparse hand-object interaction latents are appended along the temporal axis to preserve timestamp indices for self-attention propagation. We compare two alternatives that swap each axis: (i) concatenating the reference latent along the temporal axis, and (ii) concatenating sparse hand-object interaction latents along the channel axis. Both degrade all metrics. In particular, channel-wise keyframes suffer from predominantly zero-padded channels, require implicit sparse-to-dense propagation, converge slower (20k vs.\ 8k steps), and yield the worst results.

\noindent \textbf{Effectiveness of TiRoPE.} 
We replace TiRoPE with the standard 3D rotary positional encoding used in Wan2.1~\cite{wan2025}. As shown in the second row of Table~\ref{tab:ablation_compact}, Ti-SSIM drops drastically. This result indicates that without explicit temporal frequency anchoring, the model loses the ability to align interactions with their designated timestamps.

\noindent \textbf{Effectiveness of Motion Prior Injection.} 
We remove the Motion Prior Injection Module and retrain the model. The third row of Table~\ref{tab:ablation_compact} shows a clear degradation in MS-RAFT and FVD, indicating less natural motion transitions. Qualitatively, the generated objects by this variant exhibit unnatural morphing during large viewpoint changes, demonstrating the necessity of MLLM-derived semantic priors for physical plausibility. Further Q-Former design ablations and an interpretability analysis of the injected motion priors are provided in the supplementary material.

\noindent \textbf{Effectiveness of Decoupled Training Strategy.} 
We compare our two-stage training paradigm against a joint-training variant that optimizes the baseline parameters and the Motion Prior Injection Module simultaneously. Since standard quantitative metrics often fail to capture subtle foreground anomalies in video generation, we therefore validate our decoupled training strategy through qualitative analysis. As shown in Figure~\ref{fig:ablation_qual}, joint training produces conspicuous foreground artifacts. We attribute this to feature entanglement, which makes the network over-reliant on the MLLM's representations for synthesizing HOI textures. Since the MLLM is inherently prone to hallucinating visual details, injecting its produced visual features to DiT severely corrupts the object's appearance.

\subsection{Robustness to Sparse Keyframes}
\label{subsec:mpi_analysis}
We train with four keyframes and vary their number $K$ only at inference to study how the model behaves under different sparsity levels ($K$ at inference group in Table~\ref{tab:ablation_compact} ). From $K{=}2$ to $8$, FID/FVD improve while Ti-SSIM stays stable, indicating that denser sparse control improves fidelity without hurting timestamp alignment. More robustness analysis across sampling strategies in the supplementary material.

\section{Conclusion}
\label{sec:conclusion}

In this paper, we propose the SparseCtrl-HOI framework, which substitutes traditional dense frame-wise guidance with a sparse temporal control paradigm, effectively addressing the heavy annotation burden and rigid motions in Human-Object Interaction (HOI) video generation. By integrating TiRoPE for precise temporal alignment, an MLLM-driven motion prior injection, and a decoupled training strategy, our method achieves smooth transitions between interaction states and effectively prevents multimodal hallucinations. Extensive experiments demonstrate that our approach significantly outperforms existing baselines.
Ultimately, SparseCtrl-HOI provides a high-fidelity, scalable, and natural video generation solution for real-world applications such as live-streaming e-commerce.

\section*{Acknowledgements}
This work was supported by the National Natural Science Foundation of China under Grants 62476099 and 62076101, Guangdong Basic and Applied Basic Research Foundation under Grants 2024B1515020082 and 2023A1515010007, and the TCL Young Scholars Program.

\clearpage

\bibliographystyle{splncs04}
\bibliography{main}

@String(ECCV  = {Eur. Conf. Comput. Vis.})

@String(ACCV  = {Asian Conf. Comput. Vis.})

@String(ECCV  = {ECCV})

@String(ACCV  = {ACCV})

@inproceedings{HOMA,
author = {Huang, Ziyao and Zhou, Zixiang and Cao, Juan and Ma, Yifeng and Chen, Yi and Rao, Zejing and Xu, Zhiyong and Wang, Hongmei and Lin, Qin and Zhou, Yuan and Lu, Qinglin and Tang, Fan},
title = {{HOMA}: Towards Generic Human-Object Interaction in Multimodal Driven Human Animation with Weak Conditions},
year = {2025},
isbn = {9798400721373},
publisher = {Association for Computing Machinery},
address = {New York, NY, USA},
doi = {10.1145/3757377.3763861},
booktitle = {Proceedings of the SIGGRAPH Asia 2025 Conference Papers},
articleno = {187},
numpages = {12},
location = {
},
series = {SA Conference Papers '25}
}

@article{wang2025dreamactorh1,
  title={{DreamActor-H1}: High-fidelity human-product demonstration video generation via motion-designed diffusion transformers},
  author={Wang, Lizhen and Xia, Zhurong and Hu, Tianshu and Wang, Pengrui and Wei, Pengfei and Zheng, Zerong and Zhou, Ming and Zhang, Yuan and Gao, Mingyuan},
  journal={arXiv preprint arXiv:2506.10568},
  year={2025}
}

@article{zhang2025vhoi,
title = {{VHOI}: Controllable Video Generation of Human–Object Interactions from Sparse Trajectories via Motion Densification},
author = {Zhang, Wanyue and Foo, Lin Geng and Dabral, Rishabh and Beeler, Thabo and Theobalt, Christian},
journal = {arXiv preprint arXiv:2512.09646},
year = {2025},
}

@article{liu2025byteloom,
  title={{ByteLoom}: Weaving Geometry-Consistent Human-Object Interactions through Progressive Curriculum Learning},
  author={Liu, Bangya and Gong, Xinyu and Zhao, Zelin and Song, Ziyang and Lu, Yulei and Wu, Suhui and Zhang, Jun and Banerjee, Suman and Zhang, Hao},
  journal={arXiv preprint arXiv:2512.22854},
  year={2025}
}

@inproceedings{yang2023dwpose,
  title={Effective whole-body pose estimation with two-stages distillation},
  author={Yang, Zhendong and Zeng, Ailing and Yuan, Chun and Li, Yu},
  booktitle={Proceedings of the IEEE/CVF International Conference on Computer Vision},
  pages={4210--4220},
  year={2023}
}

@InProceedings{Chung16syncnet,
  author       = "Chung, J.~S. and Zisserman, A.",
  title        = "Out of time: automated lip sync in the wild",
  booktitle    = "Workshop on Multi-view Lip-reading, ACCV",
  year         = "2016",
}

@misc{qwen2.5-VL,
    title = {{Qwen2.5-VL}},
    url = {https://qwenlm.github.io/blog/qwen2.5-vl/},
    author = {Qwen Team},
    year = {2025}
}

@inproceedings{ravi2024sam2,
  title={{SAM} 2: Segment anything in images and videos},
  author={Ravi, Nikhila and Gabeur, Valentin and Hu, Yuan-Ting and Hu, Ronghang and Ryali, Chaitanya and Ma, Tengyu and Khedr, Haitham and R{\"a}dle, Roman and Rolland, Chloe and Gustafson, Laura and others},
  booktitle={International Conference on Learning Representations},
  volume={2025},
  pages={28085--28128},
  year={2025}
}

@inproceedings{zhou2023propainter,
   title={{ProPainter}: Improving Propagation and Transformer for Video Inpainting},
   author={Zhou, Shangchen and Li, Chongyi and Chan, Kelvin C.K and Loy, Chen Change},
   booktitle={Proceedings of the IEEE/CVF International Conference on Computer Vision},
   year={2023}
}

@article{wan2025,
      title={Wan: Open and Advanced Large-Scale Video Generative Models}, 
      author={Team Wan and Ang Wang and Baole Ai and Bin Wen and Chaojie Mao and Chen-Wei Xie and Di Chen and Feiwu Yu and Haiming Zhao and Jianxiao Yang and Jianyuan Zeng and Jiayu Wang and Jingfeng Zhang and Jingren Zhou and Jinkai Wang and Jixuan Chen and Kai Zhu and Kang Zhao and Keyu Yan and Lianghua Huang and Mengyang Feng and Ningyi Zhang and Pandeng Li and Pingyu Wu and Ruihang Chu and Ruili Feng and Shiwei Zhang and Siyang Sun and Tao Fang and Tianxing Wang and Tianyi Gui and Tingyu Weng and Tong Shen and Wei Lin and Wei Wang and Wei Wang and Wenmeng Zhou and Wente Wang and Wenting Shen and Wenyuan Yu and Xianzhong Shi and Xiaoming Huang and Xin Xu and Yan Kou and Yangyu Lv and Yifei Li and Yijing Liu and Yiming Wang and Yingya Zhang and Yitong Huang and Yong Li and You Wu and Yu Liu and Yulin Pan and Yun Zheng and Yuntao Hong and Yupeng Shi and Yutong Feng and Zeyinzi Jiang and Zhen Han and Zhi-Fan Wu and Ziyu Liu},
      journal = {arXiv preprint arXiv:2503.20314},
      year={2025}
}

@inproceedings{
hu2022lora,
title={Lo{RA}: Low-Rank Adaptation of Large Language Models},
author={Edward J Hu and Yelong Shen and Phillip Wallis and Zeyuan Allen-Zhu and Yuanzhi Li and Shean Wang and Lu Wang and Weizhu Chen},
booktitle={International Conference on Learning Representations},
year={2022},
url={https://openreview.net/forum?id=nZeVKeeFYf9}
}

@article{sui2025surveyhumaninteractionmotion,
  title={A survey on human interaction motion generation},
  author={Sui, Kewei and Ghosh, Anindita and Hwang, Inwoo and Zhou, Bing and Wang, Jian and Guo, Chuan},
  journal={International Journal of Computer Vision},
  volume={134},
  number={3},
  pages={113},
  year={2026},
  publisher={Springer}
}

@article{lei2024humanvideo,
      title={A Comprehensive Survey on Human Video Generation: Challenges, Methods, and Insights}, 
      author={Wentao Lei and Jinting Wang and Fengji Ma and Guanjie Huang and Li Liu},
      journal={arXiv preprint arXiv:2407.08428},
      year={2024}
}

@article{zhu2023human,
  title={Human motion generation: A survey},
  author={Zhu, Wentao and Ma, Xiaoxuan and Ro, Dongwoo and Ci, Hai and Zhang, Jinlu and Shi, Jiaxin and Gao, Feng and Tian, Qi and Wang, Yizhou},
  journal={IEEE Transactions on Pattern Analysis and Machine Intelligence},
  volume={46},
  number={4},
  pages={2430--2449},
  year={2023},
  publisher={IEEE}
}

@inproceedings{hu2025animateanyone2,
  title={Animate anyone 2: High-fidelity character image animation with environment affordance},
  author={Hu, Li and Wang, Guangyuan and Shen, Zhen and Gao, Xin and Meng, Dechao and Zhuo, Lian and Zhang, Peng and Zhang, Bang and Bo, Liefeng},
  booktitle={Proceedings of the IEEE/CVF International Conference on Computer Vision},
  pages={10207--10217},
  year={2025}
}

@inproceedings{men2025mimo,
  title={{MIMO}: Controllable character video synthesis with spatial decomposed modeling},
  author={Men, Yifang and Yao, Yuan and Cui, Miaomiao and Bo, Liefeng},
  booktitle={Proceedings of the IEEE/CVF Conference on Computer Vision and Pattern Recognition},
  pages={21181--21191},
  year={2025}
}

@inproceedings{fan2025rehold,
  title={{Re-HOLD}: Video hand object interaction reenactment via adaptive layout-instructed diffusion model},
  author={Fan, Yingying and Yang, Quanwei and Wang, Kaisiyuan and Zhou, Hang and Li, Yingying and Feng, Haocheng and Ding, Errui and Wu, Yu and Wang, Jingdong},
  booktitle={Proceedings of the IEEE/CVF Conference on Computer Vision and Pattern Recognition},
  pages={17550--17560},
  year={2025}
}

@inproceedings{pang2025manivideo,
  title={{ManiVideo}: Generating hand-object manipulation video with dexterous and generalizable grasping},
  author={Pang, Youxin and Shao, Ruizhi and Zhang, Jiajun and Tu, Hanzhang and Liu, Yun and Zhou, Boyao and Zhang, Hongwen and Liu, Yebin},
  booktitle={Proceedings of the IEEE/CVF Conference on Computer Vision and Pattern Recognition},
  pages={12209--12219},
  year={2025}
}

@article{xue2024hoiswap,
  title={{HOI-Swap}: Swapping objects in videos with hand-object interaction awareness},
  author={Xue, Zihui Sherry and Luo, Romy and Chen, Changan and Grauman, Kristen},
  journal={Advances in Neural Information Processing Systems},
  volume={37},
  pages={77132--77164},
  year={2024}
}

@InProceedings{hogue2024diffted,
    author    = {Hogue, Steven and Zhang, Chenxu and Daruger, Hamza and Tian, Yapeng and Guo, Xiaohu},
    title     = {{DiffTED}: One-shot Audio-driven {TED} Talk Video Generation with Diffusion-based Co-speech Gestures},
    booktitle = {Proceedings of the IEEE/CVF Conference on Computer Vision and Pattern Recognition Workshops},
    year      = {2024},
    pages     = {1922-1931}
}

@inproceedings{he2024s2gmmdiffusion,
  title={Co-Speech Gesture Video Generation via Motion-Decoupled Diffusion Model},
  author={He, Xu and Huang, Qiaochu and Zhang, Zhensong and Lin, Zhiwei and Wu, Zhiyong and Yang, Sicheng and Li, Minglei and Chen, Zhiyi and Xu, Songcen and Wu, Xiaofei},
  booktitle={Proceedings of the IEEE/CVF Conference on Computer Vision and Pattern Recognition},
  pages={2263--2273},
  year={2024}
}

@inproceedings{yang2025demo,
    title={Democratizing High-Fidelity Co-Speech Gesture Video Generation},
    author={Xu Yang and Shaoli Huang and Shenbo Xie and Xuelin Chen and Yifei Liu and Changxing Ding},
    booktitle={Proceedings of the IEEE/CVF International Conference on Computer Vision},
    year={2025}
}

@inproceedings{lin2025cyberhost,
  title={{CyberHost}: A One-stage Diffusion Framework for Audio-driven Talking Body Generation},
  author={Gaojie Lin and Jianwen Jiang and Chao Liang and Tianyun Zhong and Jiaqi Yang and Zerong Zheng and Yanbo Zheng},
  booktitle={International Conference on Learning Representations},
  year={2025},
  url={https://openreview.net/forum?id=vaEPihQsAA}
  }

@article{gan2025omniavatar,
      title={{OmniAvatar}: Efficient Audio-Driven Avatar Video Generation with Adaptive Body Animation}, 
      author={Qijun Gan and Ruizi Yang and Jianke Zhu and Shaofei Xue and Steven Hoi},
      journal={arXiv preprint arXiv:2506.18866},
      year={2025}
}

@article{xu2026anchorcrafter,
  title={{AnchorCrafter}: Animate Cyber-Anchors Selling Your Products via Human-Object Interacting Video Generation},
  author={Xu, Ziyi and Huang, Ziyao and Cao, Juan and Zhang, Yong and Cun, Xiaodong and Shuai, Qing and Wang, Yuchen and Bao, Linchao and Tang, Fan},
  journal={IEEE Transactions on Visualization and Computer Graphics},
  year={2026},
  publisher={IEEE}
}

@article{su2023roformerenhancedtransformerrotary,
  title={Roformer: Enhanced transformer with rotary position embedding},
  author={Su, Jianlin and Ahmed, Murtadha and Lu, Yu and Pan, Shengfeng and Bo, Wen and Liu, Yunfeng},
  journal={Neurocomputing},
  volume={568},
  pages={127063},
  year={2024},
  publisher={Elsevier}
}

@article{comanici2025gemini,
  title={Gemini 2.5: Pushing the Frontier with Advanced Reasoning, 
         Multimodality, Long Context, and Next Generation Agentic Capabilities},
  author={Comanici, Gheorghe and others},
  journal={arXiv preprint arXiv:2507.06261},
  year={2025}
}

@inproceedings{peebles2023scalable,
  title={Scalable diffusion models with transformers},
  author={Peebles, William and Xie, Saining},
  booktitle={Proceedings of the IEEE/CVF International Conference on Computer Vision},
  pages={4195--4205},
  year={2023}
}

@article{lipman2022flow,
  title={Flow matching for generative modeling},
  author={Lipman, Yaron and Chen, Ricky TQ and Ben-Hamu, Heli and Nickel, Maximilian and Le, Matt},
  journal={arXiv preprint arXiv:2210.02747},
  year={2022}
}

@misc{bilibili,
  title   = {{Bilibili}},
  author  = {{Bilibili Inc.}},
  year    = {2009},
  howpublished = {\url{https://www.bilibili.com}},
  note    = {Accessed: 2026-06-30}
}

@misc{youtube_data_api,
  title   = {{YouTube Data API v3}},
  author  = {{Google}},
  year    = {2013},
  howpublished = {\url{https://developers.google.com/youtube/v3}},
  note    = {Accessed: 2026-06-30}
}

@inproceedings{teed2020raft,
  title     = {{RAFT}: Recurrent All-Pairs Field Transforms for Optical Flow},
  author    = {Teed, Zachary and Deng, Jia},
  booktitle = {European Conference on Computer Vision (ECCV)},
  pages     = {402--419},
  year      = {2020}
}

@article{wiegand2003overview,
  title     = {Overview of the {H.264/AVC} video coding standard},
  author    = {Wiegand, Thomas and Sullivan, Gary J. and Bjontegaard, Gisle and Luthra, Ajay},
  journal   = {IEEE Transactions on Circuits and Systems for Video Technology},
  volume    = {13},
  number    = {7},
  pages     = {560--576},
  year      = {2003},
  publisher = {IEEE}
}

@misc{yt-dlp,
  title   = {{yt-dlp: A feature-rich command-line audio/video downloader}},
  author  = {{yt-dlp contributors}},
  year    = {2024},
  howpublished = {\url{https://github.com/yt-dlp/yt-dlp}},
  note    = {Accessed: 2026-06-30}
}

@misc{youtube,
  title   = {{YouTube}},
  author  = {{Google}},
  year    = {2005},
  howpublished = {\url{https://www.youtube.com}},
  note    = {Accessed: 2026-06-30}
}

@inproceedings{li2023blip,
  title={{BLIP-2}: Bootstrapping language-image pre-training with frozen image encoders and large language models},
  author={Li, Junnan and Li, Dongxu and Savarese, Silvio and Hoi, Steven},
  booktitle={International Conference on Machine Learning},
  pages={19730--19742},
  year={2023},
  organization={PMLR}
}

@inproceedings{jiang2025vace,
  title={{VACE}: All-in-one video creation and editing},
  author={Jiang, Zeyinzi and Han, Zhen and Mao, Chaojie and Zhang, Jingfeng and Pan, Yulin and Liu, Yu},
  booktitle={Proceedings of the IEEE/CVF International Conference on Computer Vision},
  pages={17191--17202},
  year={2025}
}

@inproceedings{liu2025phantom,
  title={Phantom: Subject-consistent video generation via cross-modal alignment},
  author={Liu, Lijie and Ma, Tianxiang and Li, Bingchuan and Chen, Zhuowei and Liu, Jiawei and Li, Gen and Zhou, Siyu and He, Qian and Wu, Xinglong},
  booktitle={Proceedings of the IEEE/CVF International Conference on Computer Vision},
  pages={14951--14961},
  year={2025}
}

@inproceedings{unterthiner2019fvd,
  title={{FVD}: A new metric for video generation},
  author={Unterthiner, Thomas and van Steenkiste, Sjoerd and Kurach, Karol and Marinier, Raphael and Michalski, Marcin and Gelly, Sylvain},
  booktitle={International Conference on Learning Representations},
  year={2019},
  url={https://openreview.net/forum?id=rylgEULtdN}
}

@article{sun2026coshdit,
  title={{Cosh-DiT}: Co-Speech Gesture Video Synthesis via Hybrid Audio-Visual Diffusion Transformers},
  author={Sun, Yasheng and Xu, Zhiliang and Zhou, Hang and Guan, Jiazhi and Yang, Quanwei and Wang, Kaisiyuan and Liang, Borong and Li, Yingying and Feng, Haocheng and Wang, Jingdong and others},
  journal={International Journal of Computer Vision},
  volume={134},
  number={3},
  pages={85},
  year={2026},
  publisher={Springer}
}

@inproceedings{guan2025audcast,
  title={{AudCast}: Audio-driven human video generation by cascaded diffusion transformers},
  author={Guan, Jiazhi and Wang, Kaisiyuan and Xu, Zhiliang and Yang, Quanwei and Sun, Yasheng and He, Shengyi and Liang, Borong and Cao, Yukang and Li, Yingying and Feng, Haocheng and others},
  booktitle={Proceedings of the IEEE/CVF Conference on Computer Vision and Pattern Recognition},
  pages={10678--10689},
  year={2025}
}

@inproceedings{meng2025echomimicv2,
  title={{EchoMimicV2}: Towards striking, simplified, and semi-body human animation},
  author={Meng, Rang and Zhang, Xingyu and Li, Yuming and Ma, Chenguang},
  booktitle={Proceedings of the IEEE/CVF Conference on Computer Vision and Pattern Recognition},
  pages={5489--5498},
  year={2025}
}

@inproceedings{HOIGen,
  author       = {Kun Liu and
                  Qi Liu and
                  Xinchen Liu and
                  Jie Li and
                  Yongdong Zhang and
                  Jiebo Luo and
                  Xiaodong He and
                  Wu Liu},
  title        = {{HOIGen-1M}: A Large-scale Dataset for Human-Object Interaction Video Generation},
  booktitle    = {Proceedings of the IEEE/CVF Conference on Computer Vision and Pattern Recognition},
  year         = {2025}
}

@inproceedings{huang2024vbench,
  title={{VBench}: Comprehensive benchmark suite for video generative models},
  author={Huang, Ziqi and He, Yinan and Yu, Jiashuo and Zhang, Fan and Si, Chenyang and Jiang, Yuming and Zhang, Yuanhan and Wu, Tianxing and Jin, Qingyang and Chanpaisit, Nattapol and others},
  booktitle={Proceedings of the IEEE/CVF Conference on Computer Vision and Pattern Recognition},
  pages={21807--21818},
  year={2024}
}

@InProceedings{corona2025vlogger,
    author    = {Corona, Enric and Zanfir, Andrei and Bazavan, Eduard Gabriel and Kolotouros, Nikos and Alldieck, Thiemo and Sminchisescu, Cristian},
    title     = {{VLOGGER}: Multimodal Diffusion for Embodied Avatar Synthesis},
    booktitle = {Proceedings of the IEEE/CVF Conference on Computer Vision and Pattern Recognition},
    year      = {2025},
    pages     = {15896-15908}
}

@article{liang2025alignhuman,
  title={{AlignHuman}: Improving motion and fidelity via timestep-segment preference optimization for audio-driven human animation},
  author={Liang, Chao and Jiang, Jianwen and Liao, Wang and Yang, Jiaqi and Zeng, Weihong and Liang, Han and others},
  journal={arXiv preprint arXiv:2506.11144},
  year={2025}
}

@misc{pyscenedetect,
  author = {Castellano, Brandon},
  title = {{PySceneDetect}: Intelligent video scene change detection and analysis tool},
  year = {2014},
  publisher = {GitHub},
  journal = {GitHub repository},
  howpublished = {\url{https://github.com/Breakthrough/PySceneDetect}},
  note = {Version 0.6.4} 
}

@article{emon2020novel,
  title={A Novel Nudity Detection Algorithm for Web and Mobile Application Development},
  author={Emon, Rahat Yeasin},
  journal={arXiv preprint arXiv:2006.01780},
  year={2020}
}

@article{opencv_library,
    author={Bradski, G.},
    title={The {OpenCV} Library},
    journal={Dr. Dobb's Journal of Software Tools},
    year={2000},
    note={\url{https://github.com/opencv/opencv}}
}

@article{lugaresi2019mediapipe,
  title={{MediaPipe}: A framework for building perception pipelines},
  author={Lugaresi, Camillo and Tang, Jiuqiang and Nash, Hadon and McClanahan, Chris and Uboweja, Esha and Hays, Michael and Zhang, Fan and Chang, Chuo-Ling and Yong, Ming Guang and Lee, Juhyun and others},
  journal={arXiv preprint arXiv:1906.08172},
  year={2019}
}

@inproceedings{kirillov2023segment,
  title={Segment anything},
  author={Kirillov, Alexander and Mintun, Eric and Ravi, Nikhila and Mao, Hanzi and Rolland, Chloe and Gustafson, Laura and Xiao, Tete and Whitehead, Spencer and Berg, Alexander C and Lo, Wan-Yen and others},
  booktitle={Proceedings of the IEEE/CVF International Conference on Computer Vision},
  pages={4015--4026},
  year={2023}
}

@article{wang2004ssim,
  title={Image quality assessment: from error visibility to structural similarity},
  author={Wang, Zhou and Bovik, Alan C and Sheikh, Hamid R and Simoncelli, Eero P},
  journal={IEEE transactions on image processing},
  volume={13},
  number={4},
  pages={600--612},
  year={2004},
  publisher={IEEE}
}

\clearpage


\section*{Supplementary material}

# SparseCtrl-HOI: Sparse Temporal Control for Human-Object Interaction Video Generation

## Supplementary Material

Shenbo Xie, Mingrui Cai, Xu Yang, Yifei Liu, and Changxing Ding*

South China University of Technology
eeshbx34@scut.edu.cn, chxding@scut.edu.cn

This supplementary material consists of six sections. Section 1 introduces the project demonstration, highlighting our framework and visual results. Section 2 details the user study conducted to subjectively evaluate our method. In Section 3, we present additional ablation studies on our network design. Section 4 evaluates the impact of different classifier-free guidance (CFG) scales. Section 5 gives more details of our work. Finally, Section 6 discusses the limitations of our current approach.

## 1 Project Demonstration

We provide a project demonstration to showcase the overall framework and additional qualitative results of our proposed method, which is available at https://mpi-lab.github.io/SparseCtrl-HOI.

## 2 User Study

To further validate the subjective quality of our generated results, we conduct a user study. Participants are presented with 15 sets of 5-second video clips generated by our method and competing baselines. Participants are tasked with ranking the videos within each set across two primary dimensions: **Visual Plausibility** and **Motion Diversity**. Visual Plausibility assesses the overall visual quality and faithfulness, especially in temporal fluency (the absence of flickering, distortion, or unnatural jumps), fidelity to the reference images (preservation of identity, clothing, and object shape), and spatiotemporal consistency. Motion Diversity evaluates the richness of the dynamic content, encompassing the extent of human movements and the variety of human-object interaction poses. As shown in Table 1, the results demonstrate the overall user preference and the perceptual advantages of our approach.

## 3 Ablation Studies

We conduct extended ablation experiments to validate the key architectural designs of our method, with a specific focus on the Q-former module.

* Corresponding author.

**Table 1:** User study results.

| Method | Visual Fidelity | Motion Diversity |
|---|---|---|
| Anchorcrafter | 6.7% | 7.3% |
| Omniavatar | 23.4% | 23.3% |
| VACE | 22.4% | 22.3% |
| Phantom | 17.4% | 16.6% |
| Ours | **30.1%** | **30.5%** |

**Table 2:** Quantitative ablation studies on the Q-former module. The best results are highlighted in **bold**.

| Setting | FID↓ | FVD↓ | Ti-SSIM↑ | MS-RAFT↓ |
|---|---|---|---|---|
| Alternative Architecture (MLP only) | 93.68 | 622.55 | 0.66 | 0.51 |
| Module Order (Transformer → MLP) | 97.74 | 659.92 | 0.66 | 0.51 |
| Transformer Depth (2 layers) | 91.88 | 639.64 | **0.67** | 0.50 |
| Transformer Depth (8 layers) | 92.30 | 624.58 | **0.67** | 0.50 |
| Learnable Queries ($N = 8$) | 94.19 | 631.22 | 0.65 | 0.51 |
| Learnable Queries ($N = 16$) | 92.88 | 641.42 | 0.66 | 0.48 |
| Learnable Queries ($N = 32$) | 91.76 | 658.89 | **0.67** | **0.45** |
| Learnable Queries ($N = 128$) | **86.86** | 640.50 | **0.67** | 0.49 |
| **Ours** ($N = 64$, 4 layers) | 91.28 | **607.13** | **0.67** | 0.46 |

- **Alternative Architectures:** To justify the necessity of the Q-former, we introduce a baseline where the Q-former is entirely replaced by a standard MLP block.
- **Module Order within Q-former:** We compare the performance impact of swapping the sequence of the Multi-Layer Perceptron (MLP) and the Transformer blocks within the Q-former.
- **Transformer Depth:** We investigate the effect of varying the number of Transformer layers within the Q-former (e.g., 2 layers and 8 layers) to find the optimal capacity for motion prior extraction.
- **Number of Learnable Queries:** We evaluate how the capacity of the model changes with varying numbers of learnable queries (e.g., $N = 32$ and $N = 128$), seeking the optimal balance between computational efficiency and representational power.

The quantitative comparisons for these architectural variants are summarized in Table 2. **Necessity of Q-former:** Replacing the Q-former module entirely with a simple Multi-Layer Perceptron (MLP) leads to noticeable performance degradations across all metrics. This demonstrates that a naive MLP lacks

the capacity to extract complex motion priors. The cross-attention mechanism within the Q-former is essential for effectively querying and integrating motion information. **Effect of Module Order:** Our default design (MLP → Transformer) significantly outperforms the reversed sequence (Transformer → MLP) across all metrics. When this order is reversed, the model yields suboptimal context extraction and compromised generation quality. **Effect of Transformer Depth:** We also ablate the number of Transformer layers within the Q-former. Our default depth configuration provides the optimal representational capacity. **Effect of Query Count ($N$):** The number of learnable queries $N$ acts as an information bottleneck, determining the model's capacity to extract and condense motion priors. As shown in Table 2, increasing $N$ from 8 to 64 steadily improves the Fréchet Video Distance (FVD) from 631.22 to an optimal 607.13, demonstrating significantly enhanced spatiotemporal video quality. However, further increasing $N$ to 128 degrades the FVD to 640.50. While $N = 128$ yields the best Fréchet Inception Distance (FID) at 86.86, this indicates that an excessively large query set tends to overfit to frame-level spatial details at the severe expense of temporal coherence. $N = 64$ strikes the optimal balance, achieving the absolute best overall video distribution quality (FVD), the highest structural similarity (Ti-SSIM), and highly competitive temporal motion consistency (MS-RAFT).

**Table 3:** Ablation on temporal anchoring strategies for interaction keyframes. We compare TiRoPE against three alternatives that inject the target timestamp through positional embedding, additive bias, or a learnable embedding. The best results are highlighted in **bold**.

| Temporal Anchoring | FID↓ | FVD↓ | Ti-SSIM↑ | MS-RAFT↓ |
|---|---|---|---|---|
| Concatenated keyframe positional embedding | 108.5 | 749.2 | 0.65 | 0.58 |
| Additive temporal bias | 253.3 | 1682.7 | 0.41 | 0.53 |
| Learnable temporal embedding | 210.2 | 1394.1 | 0.43 | 0.57 |
| **TiRoPE** | **89.2** | **629.6** | **0.66** | **0.45** |

**Temporal Anchoring Strategy.** We further justify our TiRoPE design by comparing it against three alternative ways of anchoring interaction keyframes to their target timestamps, as summarized in Table 3. *Concatenated keyframe positional embedding* appends a separate positional embedding of the timestamp; *Additive temporal bias* adds a timestamp-dependent bias to the attention logits; and *Learnable temporal embedding* learns an embedding for each target index. All three alternatives perform substantially worse than TiRoPE across every metric. Unlike these schemes, which require the network to additionally learn the mapping between the injected signal and the target latent frames, TiRoPE lets the keyframe token share the same temporal rotary index as its target frame (Equation 1), directly aligning them within self-attention. This also distinguishes TiRoPE from the zero-padded channel-wise injection discussed in the main text, which dilutes the keyframe signal and converges more slowly.

**Keyframe Sampling Strategy.** We further evaluate the robustness of our model to how the sparse keyframes are placed along the timeline. As shown in Table 4, random, uniform, and non-uniform (geometric-interval) sampling all yield comparable performance, indicating that our method is not sensitive to the specific keyframe placement.

**Table 4:** Ablation on keyframe sampling strategies on SparseHOI-5K.

| Sampling Strategy | FID↓ | FVD↓ | Ti-SSIM↑ | MS-RAFT↓ |
|---|---|---|---|---|
| **Random** | 89.2 | 629.6 | 0.66 | 0.45 |
| Uniform | 90.7 | 597.2 | 0.67 | 0.48 |
| Non-uniform (geometric intervals) | 95.6 | 623.0 | 0.68 | 0.48 |

## 4 Classifier-Free Guidance Scales

We conduct experiments to evaluate the impact of different Classifier-Free Guidance (CFG) scales on the generation quality. As shown in Table 5, we observe that a CFG scale of 3.0 yields the most balanced and optimal performance overall. Although setting the CFG scale to 4.5 results in a lower FVD score (598.45), it leads to subtle degradation in other metrics. Consequently, taking into account the comprehensive trade-off between visual quality and motion alignment, we empirically adopt a CFG scale of 3.0 as the default configuration for our framework.

**Table 5:** Comparisons in classifier-free guidance scales on our SparseHOI-5K testset.

| CFG scale | FID↓ | FVD↓ | Ti-SSIM↑ | MS-RAFT↓ |
|---|---|---|---|---|
| 2.5 | 94.77 | 623.06 | 0.66 | 0.47 |
| 3.0 | **91.28** | 607.13 | **0.67** | **0.46** |
| 3.5 | 93.97 | 642.45 | 0.66 | 0.47 |
| 4.0 | 94.11 | 633.57 | 0.65 | 0.49 |
| 4.5 | 92.92 | **598.45** | 0.66 | 0.51 |
| 5.0 | 92.13 | 645.63 | 0.65 | 0.50 |

## 5 Detail Description

### 5.1 Details of SparseCtrl-HOI

**Baseline Input.** As briefly formulated in Equation (2) of the main text, the input $\mathbf{z}_{\text{in}}$ comprises three primary latent components: the noisy video features

$\mathbf{z}_{\text{v}}^{\text{noise}} \in \mathbb{R}^{\frac{T}{4}\times\frac{H}{8}\times\frac{W}{8}\times C}$, the reference image feature $\mathbf{z}_{\text{ref}} \in \mathbb{R}^{1\times\frac{H}{8}\times\frac{W}{8}\times C}$, and the interaction keyframe features $\mathbf{z}_{\text{ho}} \in \mathbb{R}^{K\times\frac{H}{8}\times\frac{W}{8}\times C}$. First, to enable channel-wise concatenation with the temporally extended video features, the reference image feature $\mathbf{z}_{\text{ref}}$ is repeated $\frac{T}{4}$ times along the temporal axis. This broadcasts the identity information across all video frames, resulting in a tensor of shape $\frac{T}{4}\times\frac{H}{8}\times\frac{W}{8}\times C$. Furthermore, the keyframe features $\mathbf{z}_{\text{ho}}$ are concatenated with a sampled noise tensor $\boldsymbol{\epsilon}_{\text{ho}} \in \mathbb{R}^{K\times\frac{H}{8}\times\frac{W}{8}\times C}$ along the channel dimension. The introduction of $\boldsymbol{\epsilon}_{\text{ho}}$ serves a critical dual purpose. Structurally, it pads the channel dimension to align with the combined channel size of the video and reference features ($2C$). Conceptually, it ensures that the statistical distribution of the padded features remains consistent with that of the noisy video features ($\mathbf{z}_{\text{v}}^{\text{noise}}$), which is vital for maintaining a homogeneous input distribution across the corresponding channels. Following Equation (2), we ultimately obtain the DiT input tensor $\mathbf{z}_{\text{in}} \in \mathbb{R}^{(\frac{T}{4}+K)\times\frac{H}{8}\times\frac{W}{8}\times 2C}$.

**TiRoPE.** In standard video generation models [8] employing Diffusion Transformers (DiT), 3D Rotary Positional Embedding (RoPE) [6] is applied to the noisy video latent $\mathbf{z}_{\text{v}}^{\text{noise}}$ to inject spatiotemporal coordinates. Let the sequential temporal index of the latent video features be denoted as $t \in \{0, 1, \ldots, \frac{T}{4} - 1\}$. For an interaction keyframe designed to appear at a specific physical timestamp $\tau_i$ (in seconds), we must establish a deterministic mapping from physical time to this latent temporal index space. Given a target video frame rate of $f$ (FPS), the physical frame number corresponds to $\tau_i \cdot f$. Because the video VAE encoder [8] inherently performs temporal downsampling by a factor of 4, the corresponding target latent temporal index $t_i$ is formulated as:

$$t_i = \left\lfloor \frac{\tau_i \cdot f}{4} \right\rfloor, \tag{1}$$

where $\lfloor \cdot \rfloor$ denotes the floor operation.

In our TiRoPE formulation, we explicitly assign this calculated target index $t_i$ as the temporal positional index for the latent feature of the $i$-th interaction keyframe, $\mathbf{z}_{\text{ho}}^{i}$. Formally, if $\text{RoPE}(\mathbf{z}, t, s)$ denotes the rotary embedding function for a feature $\mathbf{z}$ at temporal index $t$ and spatial index $s$, the embedding for the interaction keyframe is computed as $\text{RoPE}(\mathbf{z}_{\text{ho}}^{i}, t_i, s_{\text{ho}})$. Here, the spatial positional index $s_{\text{ho}}$ remains mathematically unchanged to maintain spatial consistency with the interaction keyframe.

**Injection of Condensed Motion Priors.** We employ a standard cross-attention architecture within each transformer block to inject the condensed motion priors $\mathbf{C}_{\text{mllm}}$ into the DiT backbone. Specifically, the intermediate spatiotemporal latent features extracted from the preceding layers of the DiT block are linearly projected to serve as the Query ($\mathbf{Q}$). $\mathbf{C}_{\text{mllm}}$ acts as the conditioning context and is linearly projected to form both the Key ($\mathbf{K}$) and the Value ($\mathbf{V}$).

Formally, this cross-attention operation can be expressed as:

$$\text{Cross-Attention}(\mathbf{Q}, \mathbf{K}, \mathbf{V}) = \text{Softmax}\left(\frac{\mathbf{Q}\mathbf{K}^T}{\sqrt{d}}\right)\mathbf{V}, \tag{2}$$

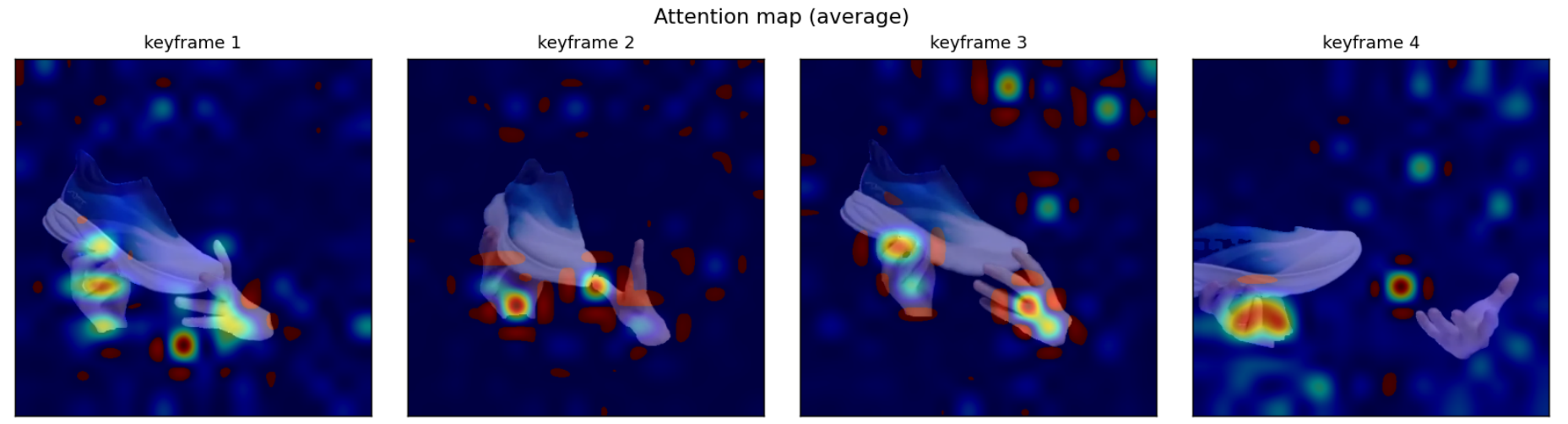


**(a)** Q-Former cross-attention averaged over the learnable queries and mapped back onto the four input keyframes.

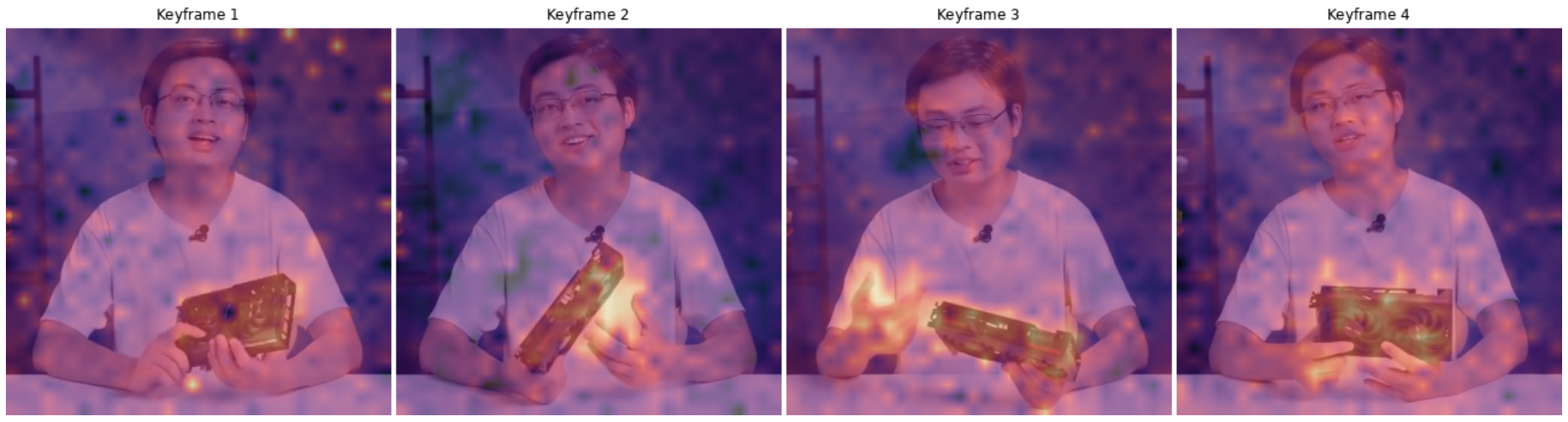


**(b)** Per-pixel feature difference at the 39th DiT layer with and without $\mathbf{C}_{\mathrm{mllm}}$, shown at the four target timestamps.

**Fig. 1:** Interpretability of the Motion Prior Injection (M.P.I.) module. Warmer colors in (a) denote higher attention; the maps in (b) are normalized to $[0, 1]$ per frame.

where $d$ represents the feature dimension of the queries and keys.

**Motion-Prior Interpretability.** Figure 1 analyzes how the Motion Prior Injection (M.P.I.) module guides generation. We average the Q-Former cross-attention over its learnable queries and map it back onto each input keyframe. As shown in Figure 1(a), the attention concentrates on the hand-object contact regions across all four keyframes, indicating that the condensed motion priors $\mathbf{C}_{\mathrm{mllm}}$ primarily summarize the interaction states rather than the global appearance. We also compute the feature difference at the 39th DiT layer between the models with and without $\mathbf{C}_{\mathrm{mllm}}$. As shown in Figure 1(b), injecting $\mathbf{C}_{\mathrm{mllm}}$ predominantly modulates the interacting hands and objects at the target timestamps. This is consistent with the FVD/MS-RAFT degradation observed when M.P.I. is removed in the main paper's ablation. The corresponding Qwen2.5-VL textual outputs are provided below.

**MLLM Textual Output.** Figure 2 shows the four interaction keyframes fed to Qwen2.5-VL [7], and the quote below lists its textual output obtained with the prompt described in the main paper. The model correctly characterizes the change of hand pose, object pose, and hand-object contact region across the keyframes. These descriptions form the high-level motion prior that is condensed into $\mathbf{C}_{\mathrm{mllm}}$ and injected into the DiT backbone.

> *"Based on the provided keyframes, here's an inferred sequence of the hand-object interaction:*

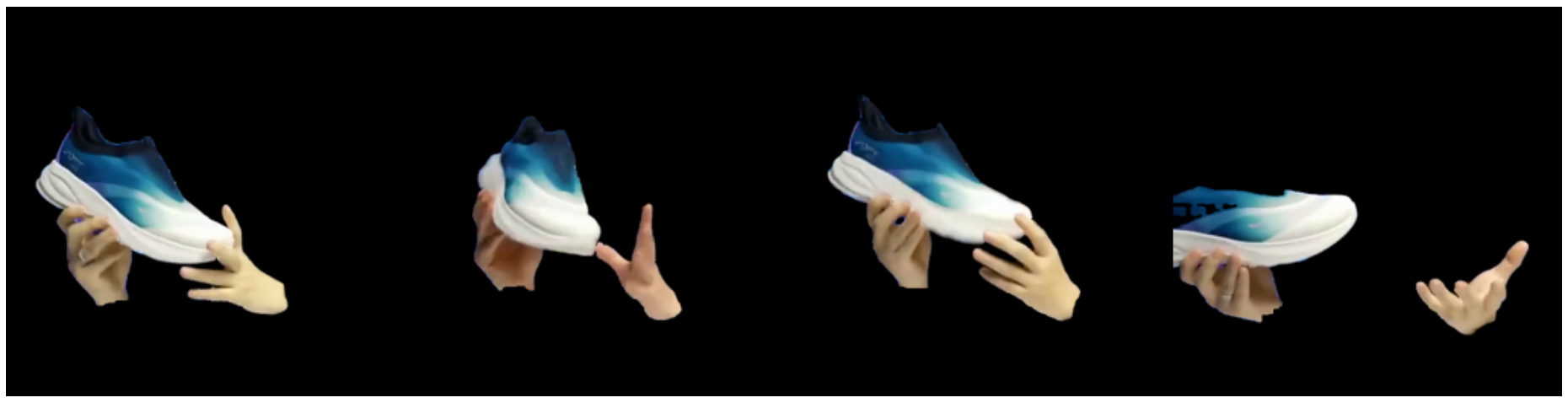

**Fig. 2:** The four input interaction keyframes used to query Qwen2.5-VL for motion-prior extraction.

*### Keyframe 1: - **Hand Pose:** The left hand is holding the shoe with the thumb and index finger gripping the side near the heel. - **Object Pose:** The shoe is positioned horizontally, with the sole facing upwards. - **Contact Region:** The thumb and index finger are in contact with the side of the shoe.*
*### Keyframe 2: - **Hand Pose:** The left hand has moved slightly, with the fingers now more spread out, and the thumb is still gripping the side of the shoe. - **Object Pose:** The shoe remains horizontal, but there is a slight rotation or tilt. - **Contact Region:** The thumb and index finger maintain contact with the side of the shoe, but the grip appears looser.*
*### Keyframe 3: - **Hand Pose:** The left hand has rotated slightly, with the fingers more spread out, and the thumb is still gripping the side of the shoe. - **Object Pose:** The shoe remains horizontal, but there is a noticeable rotation or tilt compared to Keyframe 1. - **Contact Region:** The thumb and index finger continue to be in contact with the side of the shoe, but the grip is more relaxed.*
*### Keyframe 4: - **Hand Pose:** The left hand has moved further away from the shoe, with the fingers extended and the thumb no longer gripping the side. - **Object Pose:** The shoe is now positioned vertically, with the sole facing downwards. - **Contact Region:** The hand is no longer in contact with the shoe, indicating that the interaction has ended.*
*### Summary of Hand-Object Interaction Process: 1. **Start (Keyframe 1):** The left hand holds the shoe with a firm grip near the heel. 2. **Middle (Keyframes 2 and 3):** The hand rotates slightly, maintaining a loose grip while the shoe rotates. 3. **End (Keyframe 4):** The hand moves away from the shoe, ending the interaction.*
*These keyframes suggest a smooth transition from holding the shoe to releasing it, with one hand moving away from the object. "*

### 5.2 Details on Dataset Construction

In **Phase I**, we initially compile a comprehensive list of prominent live-streamer IDs from the target e-commerce platforms [1, 2]. We then utilize automated

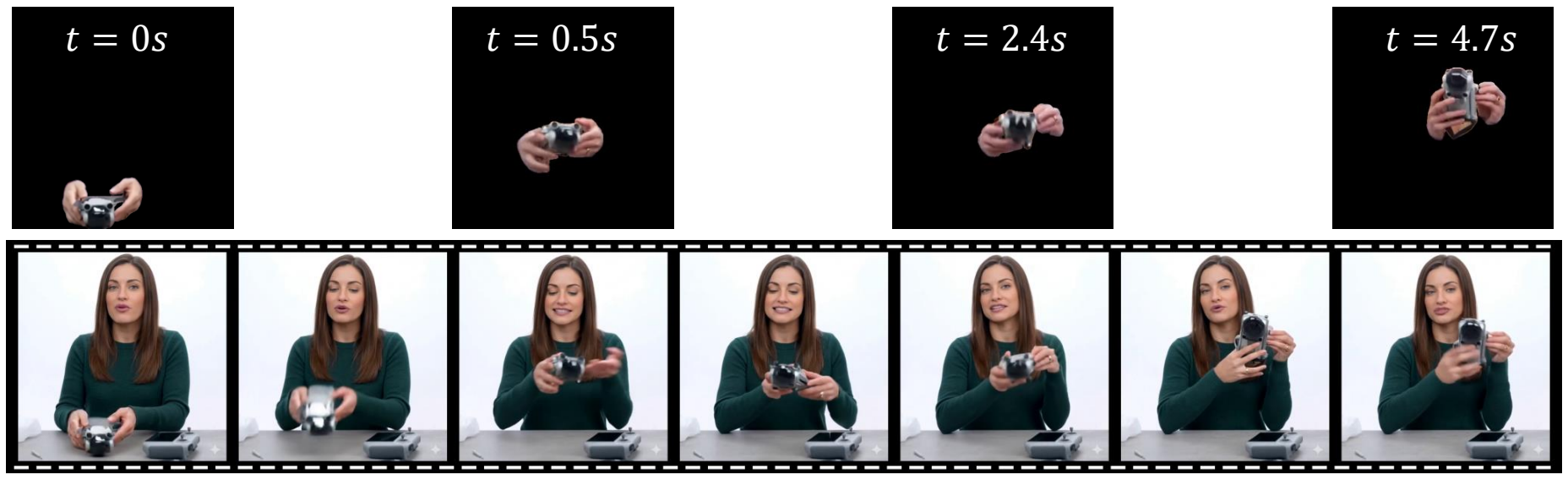


**Fig. 3:** A failure case of our work.

scripts [3, 9] to batch-download the videos from their lists. To strictly adhere to privacy regulations, portrait rights, and platform copyright policies, we will only publicly release the exact Video IDs (VIDs) alongside our automated downloading scripts. This allows the research community to independently and legally reconstruct the identical source dataset. In **Phase IV**, we design an automated, high-fidelity hand segmentation pipeline that cascades MediaPipe [5] and the Segment Anything Model (SAM) [4]. Specifically, for each video frame, we first employ MediaPipe to detect fine-grained 2D hand landmarks. We then compute a tightly bounded rectangular box around the detected hand region with a 10-pixel padding margin. This bounding box is subsequently fed as a spatial prompt into a pre-trained SAM, which predicts the pixel-level boolean mask.

### 5.3 Experimental Details

During training, we construct the reference image by randomly sampling a single frame from the inpainted, object-free video. To extract $K$ interaction keyframes, we select the first and last indices of the clip, and randomly sample the remaining $K - 2$ intermediate indices from evenly partitioned temporal windows. Finally, based on these selected temporal indices, we utilize our pre-extracted masks to crop the corresponding hand-object regions from the original RGB frames.

## 6 Limitations

While our method achieves high-quality generation in most scenarios, it still exhibits certain limitations. A primary constraint is its limited capability to handle drastic morphological changes of the interacting objects. Specifically, when an object undergoes severe shape deformations or extreme structural variations between the provided interaction keyframes, the model struggles to hallucinate a reasonable and natural spatiotemporal transition. As illustrated in Figure 3, such drastic morphological shifts can occasionally lead to visual artifacts in the generated intermediate frames. Future work will focus on enhancing the temporal prior module to better capture and render complex object state dynamics.

## References

1. Bilibili Inc.: Bilibili. https://www.bilibili.com (2009), accessed: 2026-06-30
2. Google: YouTube. https://www.youtube.com (2005), accessed: 2026-06-30
3. Google: YouTube Data API v3. https://developers.google.com/youtube/v3 (2013), accessed: 2026-06-30
4. Kirillov, A., Mintun, E., Ravi, N., Mao, H., Rolland, C., Gustafson, L., Xiao, T., Whitehead, S., Berg, A.C., Lo, W.Y., et al.: Segment anything. In: Proceedings of the IEEE/CVF International Conference on Computer Vision. pp. 4015–4026 (2023)
5. Lugaresi, C., Tang, J., Nash, H., McClanahan, C., Uboweja, E., Hays, M., Zhang, F., Chang, C.L., Yong, M.G., Lee, J., et al.: MediaPipe: A framework for building perception pipelines. arXiv preprint arXiv:1906.08172 (2019)
6. Su, J., Lu, Y., Pan, S., Murtadha, A., Wen, B., Liu, Y.: RoFormer: Enhanced transformer with rotary position embedding. arXiv preprint arXiv:2104.09864 (2023)
7. Team, Q.: Qwen2.5-VL (January 2025), https://qwenlm.github.io/blog/qwen2.5-vl/
8. Wan, T., Wang, A., Ai, B., Wen, B., Mao, C., Xie, C.W., Chen, D., Yu, F., Zhao, H., Yang, J., Zeng, J., Wang, J., Zhang, J., Zhou, J., Wang, J., Chen, J., Zhu, K., Zhao, K., Yan, K., Huang, L., Feng, M., Zhang, N., Li, P., Wu, P., Chu, R., Feng, R., Zhang, S., Sun, S., Fang, T., Wang, T., Gui, T., Weng, T., Shen, T., Lin, W., Wang, W., Wang, W., Zhou, W., Wang, W., Shen, W., Yu, W., Shi, X., Huang, X., Xu, X., Kou, Y., Lv, Y., Li, Y., Liu, Y., Wang, Y., Zhang, Y., Huang, Y., Li, Y., Wu, Y., Liu, Y., Pan, Y., Zheng, Y., Hong, Y., Shi, Y., Feng, Y., Jiang, Z., Han, Z., Wu, Z.F., Liu, Z.: Wan: Open and advanced large-scale video generative models. arXiv preprint arXiv:2503.20314 (2025)
9. yt-dlp contributors: yt-dlp: A feature-rich command-line audio/video downloader. https://github.com/yt-dlp/yt-dlp (2024), accessed: 2026-06-30



\end{document}